\documentclass[sigconf]{acmart}
\usepackage{appendix}
\usepackage{soul}
\usepackage{enumerate}
\usepackage{enumitem}
\AtBeginDocument{%
  \providecommand\BibTeX{{%
    \normalfont B\kern-0.5em{\scshape i\kern-0.25em b}\kern-0.8em\TeX}}}

\setcopyright{acmlicensed}
\copyrightyear{2024}
\acmYear{2024}
\acmDOI{10.1145/3664647.368096}
\acmConference[MM '24]{Proceedings of the 32nd ACM International Conference on Multimedia}{October 28-November 1, 2024}{Melbourne, VIC, Australia}

\acmBooktitle{Proceedings of the 32nd ACM International Conference on Multimedia (MM '24), October 28-November 1, 2024, Melbourne, VIC, Australia} 
\acmISBN{979-8-4007-0686-8/24/10}

\usepackage{algorithm}
\usepackage{algpseudocode}
\usepackage{multirow}

\begin{document}

\title{Frame Interpolation with Consecutive Brownian Bridge Diffusion}


\author{Zonglin Lyu$^1$, Ming Li$^2$, Jianbo Jiao$^3$, and Chen Chen$^2$}
\affiliation{
\institution{u1519979@umail.utah.edu, mingli@ucf.edu, j.jiao@bham.ac.uk, chen.chen@crcv.ucf.edu\\
$^1$University of Utah, $^2$Center for Research in Computer Vision, University of Central Florida, $^3$ University of Birmingham\\}
\country{\textcolor{magenta}{Project Page: zonglinl.github.io/videointerp/
}}}

%%\renewcommand{\shortauthors}{Trovato and Tobin, et al.}

\begin{abstract}
  Recent work in Video Frame Interpolation (VFI) tries to formulate VFI as a diffusion-based conditional image generation problem, synthesizing the intermediate frame given a random noise and neighboring frames. Due to the relatively high resolution of videos, Latent Diffusion Models (LDMs) are employed to run diffusion models in latent space efficiently. Such a formulation poses a crucial challenge: VFI expects that the output is \textit{deterministically} equal to the ground truth intermediate frame, but LDMs \textit{randomly} generate a diverse set of different images when the model runs multiple times. The diversity is due to the large cumulative variance (variance accumulated at each generation step) of generated latent representations in LDMs, making the sampling trajectory random. To address this problem, we propose our unique solution: Frame Interpolation with Consecutive Brownian Bridge Diffusion. Specifically, we propose consecutive Brownian Bridge diffusion that takes a deterministic initial value as input, resulting in a much smaller cumulative variance of generated latent representations. Our experiments suggest that our method can improve together with the improvement of the autoencoder and achieve state-of-the-art performance in VFI, leaving strong potential for further enhancement. Our code is available at \textcolor{magenta}{https://github.com/ZonglinL/ConsecutiveBrownianBridge}.
\end{abstract}




\begin{teaserfigure}
\centering
  \includegraphics[width=\textwidth]{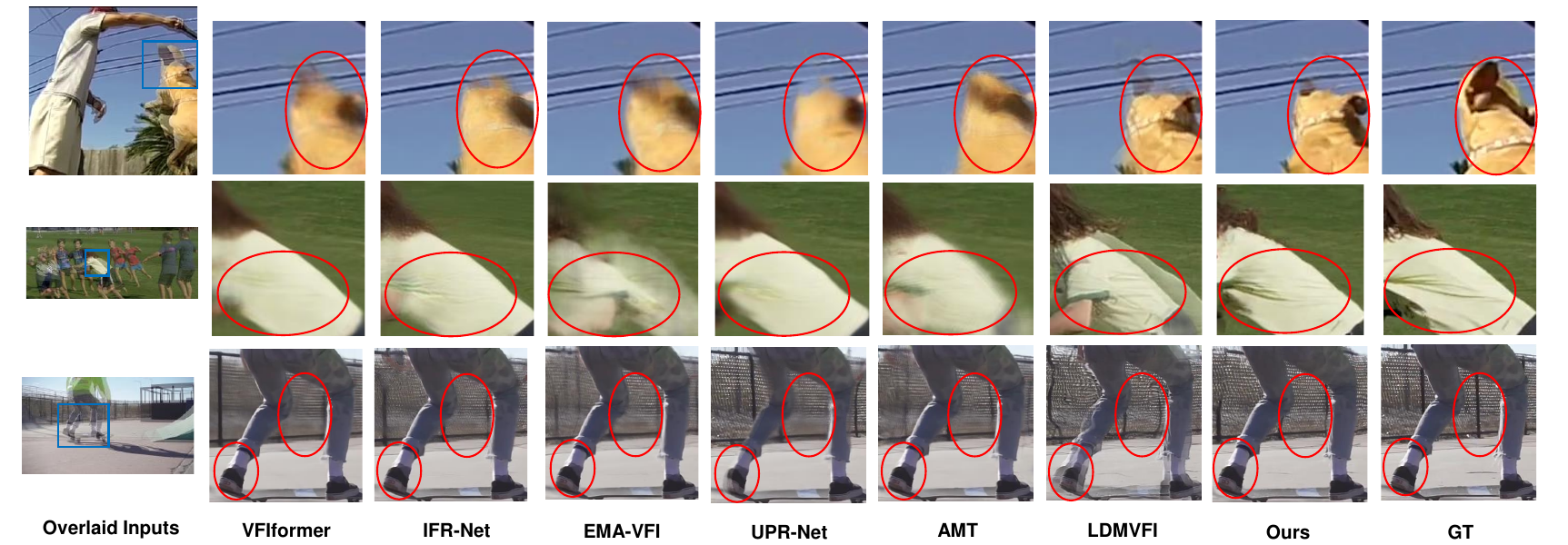}
  %\vspace{-15pt}
  \caption{Qualitative Comparison of our proposed method and SOTAs. Our method generates clear interpolated images, while recent SOTAs generate blurred or overlaid results. In our method, the generated images have clearer dog skins (first row), clearer cloth with folds (second row), and clearer fences with
    nets and high-quality shoes (third row). Images within blue boxes are displayed for detailed comparisons, and red circles highlight our performance. Examples are chosen from SNU-FILM~\cite{choi2020channel} extreme subset. More visualization results are shown in \textcolor{blue}{Appendix~\ref{appendix:qualitative}}}
  \label{fig:teaser}
\end{teaserfigure}


\maketitle

\section{Introduction}
\label{sec:intro}
Video Frame Interpolation (VFI) aims to generate high frame-per-second (fps) videos from low fps videos by estimating the intermediate frame given its neighboring frames. High-quality frame interpolation contributes to other practical applications such as novel view synthesis~\cite{flynn2016deepstereo}, video compression~\cite{wu2018video}, and high-fps cartoon synthesis~\cite{siyao2021deep}.

Current works in VFI can be divided into two folds in terms of methodologies: flow-based methods~\cite{plack2023frame,lu2022video,jin2023unified,argaw2022long,huang2022real,siyao2021deep,hu2022many,niklaus2018context,choi2021high,dutta2022non,licvpr23amt,park2023BiFormer,zhang2023extracting} and kernel-based methods~\cite{chen2021pdwn,cheng2020video,lee2020adacof,niklaus2017video,niklaus2017videosep,shi2021video,dai2017deformable}. Flow-based methods compute flows in the neighboring frames and forward warp neighboring images and features~\cite{hu2022many,niklaus2018context,niklaus2020softmax,jin2023unified,siyao2021deep} or estimate flows from the intermediate frame to neighboring frames and backward warp neighboring frames and features~\cite{lu2022video,plack2023frame,argaw2022long,huang2022real,dutta2022non,choi2021high,licvpr23amt,park2023BiFormer,zhang2023extracting}. Instead of relying on optical flows, kernel-based methods predict convolution kernels for pixels in the neighboring frames. Recent advances in flow estimation~\cite{huang2022flowformer,weinzaepfel2023croco,hui2018liteflownet,huang2022rife,ilg2017flownet,teed2020raft,sun2018pwc} make it more popular to adopt flow-based methods in VFI.

Other than these two folds of methods, MCVD~\cite{voleti2022mcvd} and LDMVFI~\cite{danier2023ldmvfi} start formulating VFI as a diffusion-based image generation problem, where LDMVFI takes advantage of LDM~\cite{rombach2022high} for better efficiency. Though diffusion models achieve excellent performance in image generation, there remain challenges in applying them to VFI. 
\begin{enumerate}[noitemsep,leftmargin=*]
    \item The formulation of diffusion models results in a large cumulative variance (the variance accumulated during sampling) of generated latent representations. The sampling process starts with standard Gaussian noise and adds small Gaussian noise to the denoised output at each step. Noises are added up to a large cumulative variance when images are generated. Though such a variance is beneficial to diversity (i.e. repeated sampling results in \textit{different} outputs), \ul{VFI requires that repeated sampling returns \textit{identical} results, which is the ground truth intermediate frame.} Therefore, a small cumulative variance is preferred in VFI. The relation of the cumulative variance and diversity is supported by the fact that DDIM~\cite{song2021denoising} tends to generate relatively deterministic images than DDPM~\cite{ho2020denoising} because DDIM removes small noises at each sampling step. LDMVFI~\cite{danier2023ldmvfi} uses conditional generation as
    guidance, but this does not change the nature of large cumulative variance. In Section~\ref{sec: consecutive BB}, we show that our method has a much lower cumulative variance.
    
    \item Videos usually have high resolution, which can be up to 4K~\cite{Perazzi_CVPR_2016}, resulting in practical constraints to apply diffusion models~\cite{ho2020denoising} in pixel spaces. It is natural to apply Latent Diffusion Models (LDMs)~\cite{rombach2022high} for better efficiency, but this does not take advantage of neighboring frames, which can be a good guide to reconstruction. LDMVFI\cite{danier2023ldmvfi} designs reconstruction models that leverage neighboring frames, but it tends to reconstruct overlaid images when there is a relatively large motion between neighboring frames, possibly due to the cross-attention with features of neighboring frames, which is shown in Figure~\ref{fig:teaser}.
\end{enumerate}

To tackle these challenges, we propose a consecutive Brownian Bridge Diffusion (in latent space) that transits among three deterministic endpoints for VFI. This method results in a much smaller cumulative variance, achieving a better estimation of the ground truth inputs. We also provide a novel method to analyze the LDM-based VFI methods: by analyzing the gap between quantitative metrics of the outputs from the autoencoder and the outputs from the diffusion model + decoder (we name this as ground truth estimation), it is easier to figure out the specific directions of improvement: whether to improve the autoencoder or the diffusion model. Moreover, we take advantage of flow estimation and refinement methods in recent literature~\cite{lu2022video} to improve the autoencoder. The feature pyramids from neighboring frames are warped based on estimated optical flows, aiming to alleviate the issues of reconstructing overlaid images. In experiments, our method improves by a large margin when the autoencoder is improved and achieves state-of-the-art performance. Our contribution can be summarized in three parts:
\begin{itemize}[noitemsep,leftmargin=*] 
    \item We propose a novel consecutive Brownian Bridge diffusion model for VFI and justify its advantages over traditional diffusion models: lower cumulative variance and better ground truth estimation capability. Additionally, we provide a cleaner formulation of Brownian Bridges and also propose the loss weighting for our Consecutive Brownian Bridges.

    \item We provide a novel method to analyze LDM-based VFI. With our methods of analysis, researchers can have specific directions for further improvements.
    
    \item Through extensive experiments, we validate the effectiveness of our method. Our method estimates the ground truth better than traditional diffusion with conditional generation (LDMVFI~\cite{danier2023ldmvfi}). Moreover, the performance of our method improves when the autoencoder improves and achieves state-of-the-art performance with a simple yet effective autoencoder, indicating its strong potential in VFI.
\end{itemize}

\section{Related Works}

\subsection{Video Frame Interpolation}
Video Frame Interpolation can be roughly divided into two categories in terms of methodologies: flow-based methods~\cite{plack2023frame,lu2022video,jin2023unified,argaw2022long,huang2022real,siyao2021deep,hu2022many,niklaus2018context,choi2021high,dutta2022non,licvpr23amt,park2023BiFormer,zhang2023extracting} and kernel-based methods~\cite{chen2021pdwn,cheng2020video,lee2020adacof,niklaus2017video,niklaus2017videosep,shi2021video,dai2017deformable}. Flow-based methods assume certain motion types, where a few works assume non-linear types~\cite{choi2021high,dutta2022non} while others assume linear. Via such assumptions, flow-based methods estimate flows in two ways. Some estimate flows from the intermediate frame to neighboring frames (or the reverse way) and apply backward warping to neighboring frames and their features~\cite{lu2022video,plack2023frame,argaw2022long,huang2022real,dutta2022non,choi2021high, licvpr23amt,park2023BiFormer,zhang2023extracting}. Others compute flows among the neighboring frames and apply forward splatting~\cite{hu2022many,niklaus2018context,niklaus2020softmax,jin2023unified,siyao2021deep}. In addition to the basic framework, advanced details such as recurrence of inputs with different resolution level~\cite{jin2023unified}, cross-frame attention~\cite{zhang2023extracting}, and 4D-correlations~\cite{licvpr23amt} are proposed to improve performance. Kernel-based methods, introduced by~\cite{niklaus2017video}, aim to predict the convolution kernel applied to neighboring frames to generate the intermediate frame, but it has difficulty in dealing with large displacement. Following works~\cite{dai2017deformable,cheng2020video,lee2020adacof} alleviate such issues by introducing deformable convolution. LDMVFI~\cite{danier2023ldmvfi} recently introduced a method based on Latent Diffusion Models (LDMs)~\cite{rombach2022high},  formulating VFI as a conditional generation task. LDMVFI uses an autoencoder introduced by LDMs to compress images into latent representations, efficiently run the diffusion process, and then reconstruct images from latent space. Instead of directly predicting image pixels during reconstruction, it takes upsampled latent representations in the autoencoder as inputs to predict convolution kernels in kernel-based methods to complete the VFI task. 

\begin{figure*}[t]
    \centering
    \includegraphics[width=\linewidth]{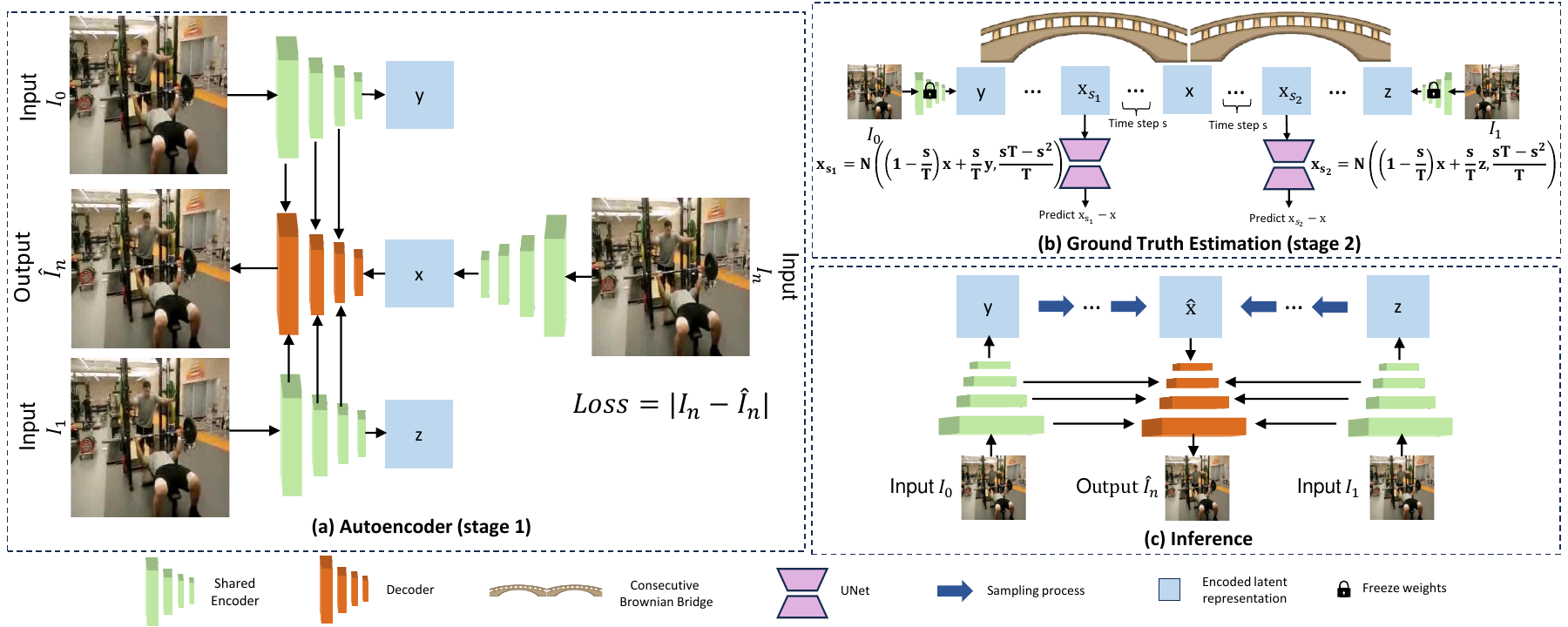}
    \caption{The illustration of our two-stage method. The encoder is shared for all frames. (a) The autoencoder stage. In this stage, previous frame $I_0$, intermediate frame $I_n$, and next frame $I_1$ are encoded by the encoder to $\mathbf{y},\mathbf{x},\mathbf{z}$ respectively. Then $\mathbf{x}$ is fed to the decoder, together with the encoder feature of $I_0,I_1$ at different down-sampling factors. The decoder predicts the intermediate frame as $\hat{I}_n$. The encoder and decoder are trained in this stage. (b) The ground truth estimation stage. In this stage, $\mathbf{y},\mathbf{x},\mathbf{z}$ will be fed to the consecutive Brownian Bridge diffusion as three endpoints, where we sample two states that move time step $s$ from $\mathbf{x}$ in both directions. The UNet predicts the difference between the current state and $\mathbf{x}$. The autoencoder is well-trained and frozen in this stage. (c) Inference. $\hat{\mathbf{x}}$ is sampled from $\mathbf{y},\mathbf{z}$ to estimate $\mathbf{x}$ (details in Section~\ref{sec: consecutive BB}). The decoder receives $\hat{\mathbf{x}}$ and encoder features of $I_0,I_1$ at different down-sampling factors to interpolate the intermediate frame.}
    \label{fig: Latent CBB}
\end{figure*}
\subsection{Diffusion Models}
The diffusion model is introduced by DDPM~\cite{ho2020denoising} to image generation task and achieves excellent performance in image generation. The whole diffusion model can be split into a forward diffusion process and a backward sampling process. The forward diffusion process is defined as a Markov Chain with steps $t = 1,...,T$, and the backward sampling process aims to estimate the distribution of the reversed Markov chain. The variance of the reversed Markov chain has a closed-form solution, and the expected value is estimated with a deep neural network. Though achieving strong performance in image generation tasks, DDPM~\cite{ho2020denoising} requires $T=1000$ iterative steps to generate images, resulting in inefficient generation. Sampling steps cannot be skipped without largely degrading performance due to its Markov chain property. To enable efficient and high-quality generation, DDIM~\cite{song2021denoising} proposes a non-Markov formulation of diffusion models, where the conditional distribution at time $t-k$ ($k>0$) can be directly computed with the conditional distribution at time $t$. Therefore, skipping steps does not largely degrade performance. Score-based SDEs~\cite{batzolis2021conditional,song2021scorebased,zhou2024denoising} are also proposed as an alternative formulation of diffusion models by writing the diffusion process in terms of Stochastic Differential Equations~\cite{oksendal2013stochastic}, where the reversed process has a closed-form continuous time formulation and can be solved with Eluer's method with a few steps~\cite{song2021scorebased}. In addition, Probability Flow ODE is proposed as the deterministic process that shares the same marginal distribution with the reversed SDE~\cite{song2021scorebased}. Following score-based SDEs, some works propose efficient methods to estimate the solution Probability Flow ODE~\cite{lu2022dpm,lu2022dpm++}. Other than using the diffusion process to connect data distribution and Gaussian distribution, diffusion bridges~\cite{li2023bbdm,zhou2024denoising,de2021diffusion,shi2024diffusion} are proposed to connect arbitrary distributions such as two different data distributions. Instead of working on the diffusion process, the Latent Diffusion Model~\cite{rombach2022high} proposes autoencoders with KL-regularized (VAE) and VQ-regularized (VQ Layer) that compress and reconstruct images, and diffusion models run with compressed images. With such autoencoders, high-resolution images can be generated efficiently. In our work, the Vector Quantized (VQ) version is deployed.

\section{Methodology}
\label{sec: Method}
In this section, we will first go through preliminaries on the Diffusion Model (DDPM)~\cite{ho2020denoising} and Brownian Bridge Diffusion Model (BBDM)~\cite{li2023bbdm} and introduce the overview of the two-stage formulation: autoencoder and ground truth estimation (with consecutive Brownian Bridge diffusion). Then, we will discuss the details of our autoencoder method. Finally, we propose our solution to the frame interpolation task: consecutive Brownian Bridge diffusion.
\begin{figure*}[t]
    \centering
    \includegraphics[width=\linewidth]{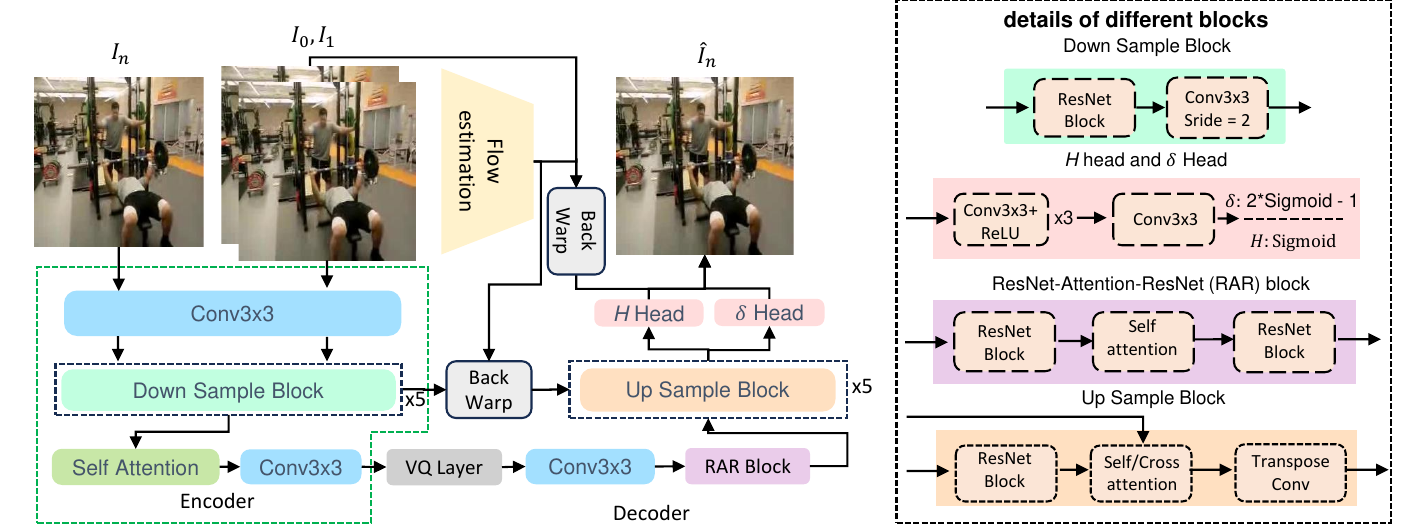}
    \caption{Architecture of the autoencoder. The encoder is in green dashed boxes, and the decoder contains all remaining parts. The output of consecutive Brownian Bridge diffusion will be fed to the VQ layer. The features of $I_0,I_1$ at different down-sampling rate will be sent to the cross-attention module at Up Sample Block in the Decoder.}
    \label{fig:autoenc}
\end{figure*}

\subsection{Preliminaries}
\textbf{Diffusion Model.} The forward diffusion process of Diffsuion Model~\cite{ho2020denoising} is defined as:
\begin{equation}
    q(\mathbf{x}_t|\mathbf{x}_{t-1}) = \mathcal{N}(\mathbf{x}_t; \sqrt{1 - \beta_t}\mathbf{x}_{t-1},\beta_t\mathbf{I}).
    \label{eq:diffusion}
\end{equation}
When $t=1$, $\mathbf{x}_{t-1} = \mathbf{x}_{0}$ is a sampled from the data (images). By iterating Eq.~\eqref{eq:diffusion}, we get the conditional marginal distribution~\cite{ho2020denoising}:
\begin{equation}
    q(\mathbf{x}_t|x_0) = \mathcal{N}(x_t;\sqrt{\alpha}_t\mathbf{x}_0,(1-\alpha_t)\mathbf{I}),
    \label{eq: ddpm marginal}
\end{equation}

\begin{equation*}
    \text{where }\alpha_t = \prod_{s=1}^t(1-\beta_s).
\end{equation*}
The sampling process is derived with the Bayes' theorem~\cite{ho2020denoising}:
\begin{equation}
    \label{eq:ddpm sample}
    p_\theta(\mathbf{x}_{t-1}|\mathbf{x}_t) = q(\mathbf{x}_{t-1}|\mathbf{x}_0,\mathbf{x}_t) = \mathcal{N}(x_{t-1};\tilde{\mathbf{\mu}}_t,\tilde{\beta}_t),
\end{equation}
\begin{equation}
    \text{where }\tilde{\mathbf{\mu}}_t = \frac{\sqrt{\alpha_{t-1}}\beta_t}{1-\alpha_t}\mathbf{x}_0 + \frac{\sqrt{1-\beta_t}(1-\alpha_{t-1})}{1-\alpha_t}\mathbf{x}_t,
    \label{eq:ddpm sample mean}
\end{equation}

\begin{equation}
     \text{and } \tilde{\beta}_t = \frac{1-\alpha_{t-1}}{1-\alpha_t}\beta_t.
\end{equation}

Eq.~\eqref{eq:ddpm sample mean} can be rewritten with Eq.~\eqref{eq: ddpm marginal} via reparameterization:

\begin{equation}
    \tilde{\mathbf{\mu}}_t = \frac{1}{1-\beta_t}\left(\mathbf{x}_t - \frac{\beta_t}{\sqrt{1-\alpha_t}}\epsilon\right) \text{, where }\epsilon\sim\mathcal{N}(0,\mathbf{I}).
    \label{eq:ddpm repara mean}
\end{equation}
By Eq.~\eqref{eq:ddpm sample mean} and~\eqref{eq:ddpm repara mean}, we only need to estimate $\epsilon$ to estimate $p_\theta(\mathbf{x}_{t-1}|\mathbf{x}_t)$. Therefore, the training objective is:
\begin{equation}
\label{eq: ddpm objective}
    \mathbb{E}_{\mathbf{x}_0,\epsilon}\left[||\epsilon_\theta(\mathbf{{x}_t},t) - \epsilon||_2^2\right].
\end{equation}
It suffices to train a neural network $\epsilon_\theta(\mathbf{x}_t,t)$ predicting $\epsilon$.

\noindent\textbf{Brownian Bridge Diffusion Model.} Brownian Bridge~\cite{ross1995stochastic} is a stochastic process that transits between two fixed endpoints, which is formulated as $X_t = W_t|(W_{t_1}, W_{t_2})$, where $W_t$ is a standard Wiener process with distribution $\mathcal{N}(0,t)$. We can write a Brownian Bridge as $X_t = W_t|(W_0, W_T)$ to define a diffusion process. When $W_0 = a,W_T = b$, we have:
\begin{equation}
    \label{eq:Brownian Bridge}
    X_t \sim\mathcal{N}\left((1-\frac{t}{T})a + \frac{t}{T}b,\frac{tT - t^2}{T}\right).
\end{equation}
BBDM~\cite{li2023bbdm} develops an image-to-image translation method based on the Brownian Bridge process by treating $a$ and $b$ as two images. The forward diffusion process is defined as:
\begin{equation}
    \label{eq:BBDM forward}
    q(\mathbf{x}_t|\mathbf{x}_0,\mathbf{y}) = \mathcal{N}\left(\mathbf{x}_t;(1-m_t)\mathbf{x}_0 + m_t\mathbf{y},\delta_t\right),
\end{equation}

\begin{equation}
    \label{eq:BBDM mt}
    \text{where } m_t = \frac{t}{T} \text{ and } \delta_t = 2s(m_t-m_t^2).
\end{equation}
$\mathbf{x}_0$ and $\mathbf{y}$ are two images, and $s$ is a constant that controls the maximum variance in the Brownian Bridge. The sampling process is derived based on Bayes' theorem~\cite{li2023bbdm}:
\begin{equation}
\begin{aligned}
\label{BBDM sample}
p_\theta(\mathbf{x}_{t-1}|\mathbf{x}_t,\mathbf{y}) &= q(\mathbf{x}_{t-1}|\mathbf{x}_0,\mathbf{x}_t,\mathbf{y})& \\ &= \frac{q(\mathbf{x}_{t}|\mathbf{x}_{t-1},\mathbf{y}) q(\mathbf{x}_{t-1}|\mathbf{x}_0,\mathbf{y})}{q(\mathbf{x}_{t}|\mathbf{x}_0,\mathbf{y}) }&\\ &= \mathcal{N}(\tilde{\mathcal{\mu}}_t,\tilde{\delta}_t\mathbf{I}).&
\end{aligned}
\end{equation}

\begin{equation*}
    \text{where } \tilde{\mathcal{\mu}}_t = c_{xt}\mathbf{x}_t + c_{yt}y + c_{\epsilon t}(m_t(\mathbf{y} - \mathbf{x}_0) + \sqrt{\delta_t}\epsilon),
\end{equation*}

\begin{equation*}
    c_{xt} = \frac{\delta_{t-1}}{\delta_t}\frac{1-m_t}{1-m_{t-1}} + \frac{\delta_{t|t-1}}{\delta_t}(1-m_t),
\end{equation*}

\begin{equation*}
    c_{yt} = m_{t-1} - m_t\frac{1-m_t}{1-m_{t-1}}\frac{\delta_{t-1}}{\delta_t},
\end{equation*}

\begin{equation*}
    c_{\epsilon t} = (1-m_{t-1})\frac{\delta_{t|t-1}}{\delta_t},
\end{equation*}

\begin{equation*}
    \delta_{t|t-1} = \delta_t - \delta_{t-1}\frac{(1-m_t)^2}{(1-m_{t-1})^2}.
\end{equation*}

It suffices to train a deep neural network $\epsilon_\theta$ to estimate the term $c_{\epsilon t}(m_t(\mathbf{y} - \mathbf{x}_0) + \sqrt{\delta_t}\epsilon)$, and therefore the training objective is $\mathbb{E}_{\mathbf{x}_0,\mathbf{y},\epsilon}[c_{\epsilon t}||m_t(\mathbf{y} -\mathbf{x}_0) + \sqrt{\delta_t}\epsilon - \epsilon_\theta(\mathbf{x_t},t)||_2^2]$.
 
\subsection{Formulation of Diffusion-based VFI}
\label{sec: formulation}
The goal of video frame interpolation is to estimate the intermediate frame $I_n$ given the previous frame $I_0$ and the next frame $I_1$. n is set to 0.5 to interpolate the frame in the middle of $I_0$ and $I_1$. In latent diffusion models~\cite{rombach2022high}, there is an autoencoder that encodes images to latent representations and decodes images from latent representations. The diffusion process denoises a latent representation, and the decoder reconstruct it back to an image. Since the initial noise is random, the decoded images are diverse images when they are sampled repetitively with the same conditions such as poses. Instead of diversity, VFI looks for a deterministic ground truth, which is the intermediate frame. To estimate the ground truth intermediate frame, we only need to estimate the corresponding latent representation in the LDM-based framework. Therefore, LDM-based VFI can be split into two stages: autoencoder and ground truth estimation. The two stages are defined as:
\begin{enumerate}[noitemsep,leftmargin=*] 
    \item \textbf{Autoencoder}. The primary function of the autoencoder is similar to image compression: compressing images to latent representations so that the diffusion model can be efficiently implemented. We denote $\mathbf{x},\mathbf{y},\mathbf{z}$ as encoded latent representations of $I_n,I_0,I_1$. In this stage, the goal is to compress $I_n$ to $\mathbf{x}$ with an encoder and then reconstruct $I_n$ from $\mathbf{x}$ with a decoder. $\mathbf{x}$ is provided to the decoder together with neighboring frames $I_0, I_1$ and their features in the encoder at different down-sampling factors. The overview of this stage is shown in Figure~\ref{fig: Latent CBB} (a). However, to interpolate the intermediate frame, $\mathbf{x}$ is unknown, so we need to estimate this ground truth.
    \item \textbf{Ground truth estimation}. In this stage, the goal is to accurately estimate $\mathbf{x}$ with a diffusion model. The diffusion model converts $\mathbf{x}$ to $\mathbf{y},\mathbf{z}$ with the diffusion process, and we train a UNet to predict the difference between the current diffusion state and $\mathbf{x}$, shown in Figure~\ref{fig: Latent CBB} (b). The sampling process of the diffusion model will convert $\mathbf{y},\mathbf{z}$ to $\mathbf{x}$ with the UNet output.
\end{enumerate}

The autoencoder is modeled with VQModel~\cite{rombach2022high} in Section~\ref{sec:autoencoder}, and the ground truth estimation is accomplished by our consecutive Brownian Bridge Diffusion in Section~\ref{sec: consecutive BB}. During inference, both stages are combined as shown in Figure~\ref{fig: Latent CBB} (c), where we decode diffusion-generated latent representation $\hat{\mathbf{x}}$. Via such formulation, we have a novel method to analyze the LDM-based VFI method. If images decoded from $\mathbf{x}$ (Figure~\ref{fig: Latent CBB} (a)) have similar visual quality to images decoded from $\hat{\mathbf{x}}$ (Figure~\ref{fig: Latent CBB} (c)), then the diffusion model achieves a strong performance in ground truth estimation, so it will be good to develop a good autoencoder. On the other way round, the performance of ground truth estimation can be potentially improved by redesigning the diffusion model.

\subsection{Autoencoder}
\label{sec:autoencoder}
Diffusion models running in pixel space are extremely inefficient in video interpolation because videos can be up to 4K in real life~\cite{Perazzi_CVPR_2016}. Therefore, we can encode images into a latent space with encoder $\mathcal{E}$ and decode images from the latent space with decoder $\mathcal{D}$. Features of $I_0,I_1$ are included because detailed information may be lost when images are encoded to latent representations~\cite{danier2023ldmvfi}. We incorporate feature pyramids of neighboring frames into the decoder stage as guidance because neighboring frames contain a large number of shared details. Given $I_n,I_0,I_1$, the encoder $\mathcal{E}$ will output encoded latent representation $\mathbf{x},\mathbf{y},\mathbf{z}$ for diffusion models and feature pyramids of $I_0,I_1$ in different down-sampling rates, denoted $\{f_y^k\},\{f_z^k\}$, where $k$ is down-sampling factor. When $k=1$, $\{f_y^k\}\text{ and }\{f_z^k\}$ represent original images. The decoder $\mathcal{D}$ will take sampled latent representation $\hat{\mathbf{x}}$ (output of diffusion model that estimates $\mathbf{x}$) and feature pyramids $\{f_y^k\},\{f_z^k\}$ to reconstruct $I_n$. In lines of equations, we have:
\begin{equation}
    \label{eq:latent BB}
    \begin{aligned}
        &\mathbf{x},\mathbf{y},\{f_y^k\},\mathbf{z},\{f_z^k\} = \mathcal{E}(I_n,I_0,I_1),&\\ & \hat{I}_n = \mathcal{D}\left(\mathbf{x},\{f_y^k\},\{f_z^k\}\right).&
    \end{aligned}
\end{equation}
Our encoder shares an identical structure with that in LDMVFI~\cite{danier2023ldmvfi}, and we slightly modify the decoder to better fit the VFI task.

\noindent\textbf{Decoding with Warped Features.} LDMVFI~\cite{danier2023ldmvfi} apply cross-attention~\cite{vaswani2017attention} to up-sampled $\mathbf{\hat{x}}$ and $f_x^k,f_y^k$. However, this does not explicitly deal with motion changes, and therefore LDMVFI usually produces overlaid results as shown in Figure~\ref{fig:teaser}. Therefore, we estimate optical flows from $I_n$ to $I_0,I_1$ and apply backward warping to the feature pyramids to tackle this problem. Suppose $\hat{x}$ is generated by our consecutive Brownian Bridge diffusion, and it is up-sampled to $h^k$ where $k$ denotes the down-sampling factor compared to the original image. Then, we apply $CA\left(h^k,Cat(w(f_y^k),w(f_z^k))\right)$ for $k>1$ to fuse the latent representation $h^k$ and feature pyramids $f_y^k$ and $f_z^k$, where $CA(\cdot,\cdot)$, $Cat(\cdot,\cdot)$, and $w(\cdot)$ denotes cross attention, channel-wise concatenation, and backward warping with estimated optical flows respectively. Finally, we apply convolution layers to $h^1$ to predict soft mask $H$ and residual $\delta$. The interpolation output is $\hat{I}_n = H*w(I_0) + (1-H)*w(I_1) + \delta$, where $*$ holds for Hadamard product, and $\hat{I}_n$ is the reconstructed image. The detailed illustration of the architecture is shown in Figure~\ref{fig:autoenc}. The VQ layer is connected with the encoder during training, but it is disconnected from the encoder and receives the sampled latent representation from the diffusion model.

\subsection{Consecutive Brownian Bridge Diffusion}
\label{sec: consecutive BB}

Brownian Bridge diffusion model (BBDM)~\cite{li2023bbdm} is designed for translation between image pairs, connecting two deterministic points, which seems to be a good solution to estimate the ground truth intermediate frame. However, it does not fit the VFI task. In VFI, images are provided as triplets because we aim to reconstruct intermediate frames giving neighboring frames, resulting in three points that need to be connected. If we construct a Brownian Bridge between the intermediate frame and the next frame, then the previous frame is ignored, and so is the other way round. This is problematic because we do not know what "intermediate" is if we lose one of its neighbors. Therefore, we need a process that transits among three images. Given two neighboring images $I_0, I_1$, we aim to construct a Brownian Bridge process with endpoints $I_0, I_1$ and additionally condition its middle stage on the intermediate frame $I_{n}$ ($n=0.5$ for $2\times$ interpolation). To achieve this, the process starts at $t=0$ with value $\mathbf{y}$, passes $t = T$ with value $\mathbf{x}$, and ends at $t=2T$ with value $\mathbf{z}$. To be consistent with the notation in diffusion models, $\mathbf{x},\mathbf{y},\mathbf{z}$ are used to represent latent representations of $I_n, I_0, I_1$ respectively. It is therefore defined as $X_t = W_t|W_0 = \mathbf{y},W_T = \mathbf{x},W_{2T} = \mathbf{z}$. The sampling process starts from time $0$ and $2T$ and goes to time $T$. Such a process indeed consists of two Brownian Bridges, where the first one ends at $\mathbf{x}$ and the second one starts at $\mathbf{x}$. We can easily verify that for $0<t<h$:

\begin{equation}
\label{eq: split BB}
    W_s|(W_0,W_t,W_h) = 
    \begin{cases}
        W_s|(W_0,W_t) & \text{if $s<t$}\\
        W_s|(W_t,W_h) & \text{if $s>t$}
    \end{cases}.
\end{equation}

According to Eq.~\eqref{eq: split BB}, we can derive the distribution of our consecutive Brownian Bridge diffusion (details shown in \textcolor{blue}{Appendix~\ref{appendix:formula}}):

\begin{equation}
\label{eq: consecutive BB forward}
\resizebox{0.4\textwidth}{!}{$
  q(\mathbf{x}_t|\mathbf{y},\mathbf{x},\mathbf{z}) =
    \begin{cases}
      \mathcal{N}(\frac{s}{T}\mathbf{x} + (1-\frac{s}{T})\mathbf{y},\frac{s(T-s)}{T}\mathbf{I}) \text{ $s = T-t$, $t < T$}\\
      \mathcal{N}(\frac{s}{T}\mathbf{x} + (1-\frac{s}{T})\mathbf{z},\frac{s(T-s)}{T}\mathbf{I}) \text{ $s = t-T$, $t > T$}
    \end{cases}.$
    }
\end{equation}

\begin{algorithm}[t]
    \caption{Training}
    \label{alg:train}
    \begin{algorithmic}[1]
        \Repeat 
        \State sample triplet $\mathbf{x,y,z}$ from dataset
        \State $s \gets Uniform(0,T)$
        \State $w_s \gets min\{\frac{1}{\delta_t},\gamma\}$ \Comment{$\gamma$ is a pre-defined constant}
        \State $\epsilon \gets \mathcal{N}(\mathbf{0},\mathbf{I})$
        \State $\mathbf{x_{s_1}} \gets  \frac{s}{T}\mathbf{x} + (1-\frac{s}{T})\mathbf{y} + \sqrt{\frac{s(T-s)}{T}}\epsilon$
        \State $\mathbf{x_{s_2}} \gets  \frac{s}{T}\mathbf{x} + (1-\frac{s}{T})\mathbf{z} + \sqrt{\frac{s(T-s)}{T}}\epsilon$
        \State r $\gets Uniform(0,1)$
        \If{r < 0.5} take a gradient step on
        \State $\nabla_\theta||\epsilon_\theta(\mathbf{x}_{s_1},T-s,\mathbf{y},\mathbf{z}) - (\mathbf{x}_{s_1} - \mathbf{x})||_2^2$
        \Else \text{ }take a gradient step on
        \State $\nabla_\theta||\epsilon_\theta(\mathbf{x}_{s_2},T+s,\mathbf{y},\mathbf{z}) - (\mathbf{x}_{s_2} - \mathbf{x})||_2^2$
        \EndIf
        \Until{convergence}
    \end{algorithmic}
\end{algorithm}

\begin{algorithm}[t]
    \caption{Sampling}
    \label{alg:sample}
    \begin{algorithmic}[1]
        \State $t_1,t_2 \gets T, \Delta_t \gets \frac{T}{\text{sampling steps}}, \mathbf{x}_{T_1} = \mathbf{y}, \mathbf{x}_{T_2} = \mathbf{z}$
        \Repeat 
        \State $s_1,s_2 \gets t_1- \Delta_t,t_2 - \Delta_t$
        \State $\epsilon \gets \mathcal{N}(\mathbf{0},\mathbf{I})$
        \State $\mathbf{x_{s_1}} \gets x_{t_1} - \frac{\Delta_t}{t_1}\epsilon_\theta(x_{t_1},T-t_1,\mathbf{y},\mathbf{z}) + \sqrt{\frac{s_1\Delta_t}{t_1}}\epsilon$
        \State $\mathbf{x_{s_2}} \gets x_{t_2} - \frac{\Delta_t}{t_2}\epsilon_\theta(x_{t_2},T-t_2,\mathbf{y},\mathbf{z}) + \sqrt{\frac{s_2\Delta_t}{t_2}}\epsilon$
        \State $t_1,t_2 \gets s_1,s_2$

        \Until{$t_1,t_2= 0$}
    \end{algorithmic}
\end{algorithm}

\begin{table*}[t]
    \centering
    \caption{Quantitative results (LPIPS/FloLPIPS/FID, the lower the better) on test datasets. \textcolor{red}{$\dagger$} means we evaluate our consecutive Brownian Bridge diffusion (trained on Vimeo 90K triplets~\cite{xue2019video}) with autoencoder provided by LDMVFI~\cite{danier2023ldmvfi}. The best performances are \textbf{boldfaced}, and the second best performances are \underline{underlined}.}
    \label{tab:results}
    \resizebox{\textwidth}{!}{
        \begin{tabular}{lccccccc}
        \toprule
       \multirow{2}{*}{Methods} &\multirow{2}{*}{Middlebury} & \multirow{2}{*}{UCF-101}& \multirow{2}{*}{DAVIS}& \multicolumn{4}{c}{SNU-FILM} \\
        \cmidrule{5-8}
        & & & &easy & medium & hard&extreme\\
        & LPIPS/FloLPIPS/FID&LPIPS/FloLPIPS/FID&LPIPS/FloLPIPS/FID&LPIPS/FloLPIPS/FID&LPIPS/FloLPIPS/FID&LPIPS/FloLPIPS/FID&LPIPS/FloLPIPS/FID\\
        \midrule
        ABME'21~\cite{park2021ABME} &0.027/0.040/11.393 &0.058/0.069/37.066 &0.151/0.209/16.931 & 0.022/0.034/6.363 & 0.042/0.076/15.159 & 0.092/0.168/34.236&0.182/0.300/63.561\\
        
        MCVD'22~\cite{voleti2022mcvd} &0.123/0.138/41.053 &0.155/0.169/102.054& 0.247/0.293/28.002 &  0.199/0.230/32.246 & 0.213/0.243/37.474 &0.250/0.292/51.529&0.320/0.385/83.156\\

        VFIformer'22~\cite{lu2022video} &  0.015/0.024/9.439 & 0.033/0.040/22.513 & 0.127/0.184/14.407 &0.018/0.029/5.918 &0.033/0.053/11.271 &0.061/0.100/22.775 &0.119/0.185/40.586 \\
        
        IFRNet'22~\cite{Kong_2022_CVPR} &0.015/0.030/10.029&0.029/0.034/20.589&0.106/0.156/\underline{12.422} & 0.021/0.031/6.863&0.034/0.050/12.197 &0.059/0.093/23.254 &0.116/0.182/42.824 \\

        AMT'23~\cite{licvpr23amt} &0.015/0.023/\underline{7.895}&0.032/0.039/21.915 & 0.109/0.145/13.018 &0.022/0.034/6.139 & 0.035/0.055/11.039 & 0.060/0.092/20.810&0.112/0.177/40.075 \\
        
        UPR-Net'23~\cite{jin2023unified} & 0.015/0.024/7.935 &0.032/0.039/21.970 & 0.134/0.172/15.002 & 0.018/0.029/\underline{5.669}&0.034/0.052/\ul{10.983}  &0.062/0.097/22.127 & 0.112/0.176/40.098 \\
        EMA-VFI'23~\cite{zhang2023extracting} & 0.015/0.025/8.358 & 0.032/0.038/21.395& 0.132/0.166/15.186 & 0.019/0.038/5.882 & 0.033/0.053/11.051 & 0.060/0.091/\underline{20.679}& 0.114/0.170/\underline{39.051}\\
        
        LDMVFI'24~\cite{danier2023ldmvfi} & 0.019/0.044/16.167 &\underline{0.026}/0.035/26.301 & 0.107 0.153/12.554 &0.014/0.024/5.752 & \underline{0.028}/0.053/12.485 & 0.060/0.114/26.520& 0.123/0.204/47.042\\
        \midrule
        Ours$\dagger$ &\underline{0.017}/0.040/14.447  &\underline{0.024}/\underline{0.034}/\underline{15.335} &\underline{0.102}/\underline{0.150}/12.623 &\underline{0.013}/\underline{0.022}/5.737  &\underline{0.028}/\underline{0.050}/12.569  &\underline{0.058}/0.110/25.567 &0.118/0.197/46.088 \\
        Ours &\textbf{0.009}/\textbf{0.018}/\textbf{7.470} &\textbf{0.021}/\textbf{0.032}/\textbf{14.000} & \textbf{0.092}/\textbf{0.136}/\textbf{9.220}&\textbf{0.012}/\textbf{0.019}/\textbf{4.791} &\textbf{0.022}/\textbf{0.039}/\textbf{9.039} &\textbf{0.047}/\textbf{0.091}/\textbf{18.589} &\textbf{0.104}/0.184/\textbf{36.631}\\
        \bottomrule
    \end{tabular}
  }
\end{table*}

\noindent\textbf{Cleaner Formulation.} Eq.~\eqref{BBDM sample} is in a discrete setup, and the sampling process is derived via Bayes' theorem, resulting in a complicated formulation. To preserve the maximum variance, it suffices to have  $T = 2s$ in Eq.~\eqref{eq:Brownian Bridge} with a continuous formulation and discretize it for training and sampling. Our forward diffusion is defined as Eq.~\eqref{eq: consecutive BB forward}. To sample at time $s$ from $t$ ($s<t$), we rewrite Eq.~\eqref{BBDM sample} according to Eq.~\eqref{eq: split BB}:

\begin{equation}
    \label{eq:consecutive BB back}
    \begin{aligned}
    p_\theta(\mathbf{x}_{s}|\mathbf{x}_t,\mathbf{y}) &= q(\mathbf{x}_{s}|\mathbf{x},\mathbf{x}_t,\mathbf{y}) = q(\mathbf{x}_{s}|\mathbf{x},\mathbf{x}_t)& \\  &= \mathcal{N}\left(\mathbf{x}_s;\frac{s}{t}\mathbf{x}_t + (1 - \frac{s}{t})\mathbf{x},\frac{s(t-s)}{t}\mathbf{I}\right)& \\ & = \mathcal{N}\left(\mathbf{x}_s;\mathbf{x}_t - \frac{t-s}{t}(\mathbf{x}_t - \mathbf{x}),\frac{s(t-s)}{t}\mathbf{I}\right).&
    \end{aligned}
\end{equation}
Note that $\mathbf{x}_0$ in Eq.~\eqref{BBDM sample} and $\mathbf{x}$ in our formulation both represent the image. This formulation can be solved with a few steps without DDIM~\cite{song2021denoising} similar to Euler's method.

\noindent\textbf{Training and Sampling.} According to Eq.~\eqref{eq:consecutive BB back}, it suffices to have a neural network $\epsilon_\theta$ estimating $\mathbf{x}_t - \mathbf{x}_0$. Moreover, based on Eq.~\eqref{eq: consecutive BB forward}, we can sample $s$ from $Uniform(0,T)$ and compute $t = T \pm s$ for $t > T \text{ and } T <t$. With one sample of $s$, we can obtain two samples at each side of our consecutive Brownian bridge diffusion symmetric at T. $\mathbf{y},\mathbf{z}$ are added to the denoising UNet as extra conditions. Therefore, the training objective becomes:

\begin{equation}
    \label{eq: training obj}
    \begin{aligned}
        &\mathbb{E}_{\{\mathbf{y},\mathbf{x},\mathbf{z}\},\epsilon}[||\epsilon_\theta(\mathbf{x}_{s_1},T-s,\mathbf{y},\mathbf{z}) - (\mathbf{x}_{s_1} - \mathbf{x})||_2^2]& \\ &+ \mathbb{E}_{\{\mathbf{y},\mathbf{x},\mathbf{z}\},\epsilon}[||\epsilon_\theta(\mathbf{x}_{s_2},T+s,\mathbf{y},\mathbf{z}) - (\mathbf{x}_{s_2} - \mathbf{x})||_2^2].&
    \end{aligned}
\end{equation}

\begin{equation}
    \begin{aligned}
    \text{where }&\mathbf{x_{s_1}} = \frac{s}{T}\mathbf{x} + (1-\frac{s}{T})\mathbf{y} + \sqrt{\frac{s(T-s)}{T}}\epsilon,&\\& \mathbf{x_{s_2}} = \frac{s}{T}\mathbf{x} + (1-\frac{s}{T})\mathbf{z} + \sqrt{\frac{s(T-s)}{T}}\epsilon,& \\ &\epsilon \sim \mathcal{N}(\mathbf{0},\mathbf{I}).&
    \end{aligned}
\end{equation}

Optimizing Eq.~\eqref{eq: training obj} requires two forward calls of UNet. For efficiency, we randomly select one of them to optimize during training. Moreover, \cite{hang2023efficient} proposes $min-SNR-\gamma$ loss weighting for different time steps based on the signal-to-noise ratio, defined as $min\{SNR(t),\gamma\}$. In DDPM~\cite{ho2020denoising}, we have $SNR(t) = \frac{\alpha_t}{1-\alpha_t}$ because the mean and standard deviation are scaled by $\sqrt{\alpha_t}$ and $\sqrt{1-\alpha_t}$ respectively in the diffusion process. In our formulation, the expected values are not scaled down: neighboring frames share almost identical expected values. Therefore, the SNR is defined as $\frac{1}{\delta_t}$, where $\delta_t$ is the standard deviation of the diffusion process at time $t$. The weighting is defined as $w_t = min\{\frac{1}{\delta_t},\gamma\}$. 

The training algorithm is shown in Algorithm~\ref{alg:train}. To sample from neighboring frames, we sample from either of the two endpoints $\mathbf{y},\mathbf{z}$ with Eq.~\eqref{eq: consecutive BB forward} and~\eqref{eq:consecutive BB back}, shown in Algorithm~\ref{alg:sample}. After sampling, we replace $\mathbf{x}$ in Eq~\eqref{eq:latent BB} with the sampled latent representations to decode the interpolated frame.

\noindent\textbf{Cumulative Variance}. As we claimed, diffusion model~\cite{ho2020denoising} with conditional generation has a large cumulative variance while ours is much smaller. The cumulative variance for traditional conditional generation is larger than $1 + \sum_t\hat{\beta}_t$, which corresponds to 11.036 in experiments. However, in our method, such a cumulative variance is smaller than $T = 2$ in our experiments, resulting in a more deterministic estimation of the ground truth latent representations. Detailed justification is shown in \textcolor{blue}{Appendix~\ref{appendix: cum var}}

\section{Experiments}
\subsection{Implementations}

\textbf{Autoencoder.} The down-sampling factor is set to be $f=32$ for our autoencoder, which follows the setup of LDMVFI~\cite{danier2023ldmvfi}. The flow estimation and refinement modules are initialized from pretrained VFIformer~\cite{lu2022video} and frozen for better efficiency. The codebook size and embedding dimension of the VQ Layer are set to 16384 and 3 respectively. The number of channels in the latent space (encoder output) is set to 8. A self-attention~\cite{vaswani2017attention} is applied at $32\times$ down-sampling latent representation (both encoder and decoder), and cross attentions~\cite{vaswani2017attention} with warped features are applied on the $2\times$ to $32\times$ down-sampling factors in the decoder. Following LDMVFI, max-attention~\cite{tu2022maxvit} is applied for better efficiency. The model is trained with Adam optimizer~\cite{Adam} with a learning rate of $10^{-5}$ for 100 epochs with a batch size of 16.

\noindent\textbf{Consecutive Brownian Bridge Diffusion.} We set $T = 2$ (corresponding to maximum variance $\frac{1}{2}$) and discretize 1000 steps for training and 50 steps for sampling. The denoising UNet takes the concatenation of $\mathbf{x}_t,\mathbf{y},\mathbf{z}$ as input and is trained with Adam optimizer~\cite{Adam} with $10^{-4}$ learning rate for 30 epochs with a batch size of 64. $\gamma$ is set to be 5 in the $min-SNR-\gamma$ weighting.

\begin{table*}[t]
    \centering
    \caption{Ablation studies of autoencoder and ground truth estimation. + GT means we input ground truth x to the decoder part of autoencoder. + BB indicates our consecutive Brownian Bridge diffusion trained with autoencoder of LDMVFI. With our
    consecutive Brownian Bridge diffusion, the interpolated frame has almost the same performance as the interpolated frame
    with ground truth latent representation, indicating the strong ground truth estimation capability Our autoencoder also has better performance than LDMVFI~\cite{danier2023ldmvfi}.}
    \label{tab:ablation}
    \resizebox{\textwidth}{!}{
        \begin{tabular}{lccccccc}
        \toprule
        \multirow{2}{*}{Methods}&\multirow{2}{*}{Middlebury} & \multirow{2}{*}{UCF-101}& \multirow{2}{*}{DAVIS}& \multicolumn{4}{c}{SNU-FILM} \\
        \cmidrule{5-8}
        & & & &easy & medium & hard&extreme\\
        &LPIPS/FloLPIPS/FID&LPIPS/FloLPIPS/FID&LPIPS/FloLPIPS/FID&LPIPS/FloLPIPS/FID&LPIPS/FloLPIPS/FID&LPIPS/FloLPIPS/FID&LPIPS/FloLPIPS/FID\\
        \midrule
       LDMVFI'24~\cite{danier2023ldmvfi} & 0.019/0.044/16.167 &0.026/0.035/26.301 & 0.107 0.153/12.554 & 0.014/0.024/5.752 & 0.028/0.053/12.485 & 0.060/0.114/26.520& 0.123 0.204/47.042\\

       LDMVFI'24~\cite{danier2023ldmvfi} + BB &0.017/0.040/14.447 &0.024/0.034/15.335 &0.102/0.150/12.623 &0.013/0.022/5.737  &0.028/0.050/12.569  &0.058/0.110/25.567 &0.118/0.197/46.088 \\

       LDMVFI'24~\cite{danier2023ldmvfi} + GT &0.017/0.040/14.447  &0.024/0.034/15.335 &0.102/0.150/12.625 &0.013/0.022/5.739  &0.028/0.050/12.563  &0.058/0.110/25.565 &0.118/0.197/46.080 \\
        
       Ours &0.009/0.018/7.470 &0.021/0.032/14.000 &0.092/0.0136/9.220 &0.012/0.019/4.791 &0.022/0.039/9.039 &0.047/0.091/18.589 & 0.104/0.184/36.631 \\

       Ours + GT &0.009/0.018/7.468 &0.021/0.032/14.000 &0.092/0.136/9.220 &0.012/0.019/4.791 & 0.022/0.039/9.039& 0.047/0.091/18.591&0.104/0.184/36.633 \\
        
        \bottomrule
    \end{tabular}
  }
\end{table*}

\subsection{Datasets and Evaluation Metrics}
\textbf{Training Sets.} To ensure a fair comparison with most recent works~\cite{plack2023frame,lu2022video,jin2023unified,argaw2022long,huang2022real,siyao2021deep,hu2022many,niklaus2018context,choi2021high,dutta2022non}, we train our models in Vimeo 90K triplets dataset~\cite{xue2019video}, which contains 51,312 triplets. We apply random flipping, random cropping to $256\times256$, temporal order reversing, and random rotation with multiples of 90 degrees as data augmentation.

\noindent\textbf{Test Sets.} We select UCF-101~\cite{soomro2012ucf101}, DAVIS~\cite{Perazzi_CVPR_2016}, SNU-FILM~\cite{choi2020channel}, and Middlebury~\cite{baker2011database} to evaluate our method. UCF-101 and Middlebury consist of relatively low-resolution videos (less than 1K), whereas DAVIS and SNU-FILM consist of relatively high-resolution videos (up to 4K). SNU-FILM consists of four categories with increasing levels of difficulties (i.e. larger motion changes): easy, medium, hard, and extreme.

\noindent\textbf{Evaluation Metrics.} Recent works~\cite{danier2023ldmvfi,danier2022flolpips,zhang2018perceptual} reveal that PSNR and SSIM~\cite{wang2004image} are sometimes unreliable because they have relatively lower correlation with humans' visual judgments. However, learning-based metrics such as FID~\cite{heusel2017gans}, LPIPS~\cite{zhang2018perceptual}, and FloLPIPS~\cite{danier2022flolpips} are shown to have a higher correlation with humans' visual judgments in~\cite{danier2023ldmvfi,zhang2018perceptual}. Moreover, we also experimentally find such inconsistencies between PSNR/SSIM and visual quality, which will be discussed in Section~\ref{sec: exp result}. Therefore, we select FID, LPIPS, and FloLPIPS as our main evaluation metrics. LPIPS and FID measure similarities or distances in the latent space of deep learning models. FloLPIPS is based on LPIPS but takes the motion change among three frames into consideration. The results in
PSNR/SSIM are included in \textcolor{blue}{Appendix~\ref{appendix:psnr}}.

\subsection{Experimental Results}
\label{sec: exp result}

\textbf{Quantitative Results.} Our method is compared with recent open-source state-of-the-art VFI methods, such as ABME~\cite{park2021ABME}, MCVD~\cite{voleti2022mcvd}, VFIformer~\cite{lu2022video}, IFRNet~\cite{Kong_2022_CVPR}, AMT~\cite{licvpr23amt}, UPR-Net~\cite{jin2023unified}, EMA-VFI~\cite{zhang2023extracting}, and LDMVFI~\cite{danier2023ldmvfi}. The evaluation is reported in LPIPS/FloLPIPS/FID (lower the better), shown in Table~\ref{tab:results}. We evaluate VFIformer, IFRNet, AMT, UPR-Net, and EMA-VFI with their provided weights, and other results are from the appendix of LDMVFI~\cite{danier2023ldmvfi}. Models with different versions in the number of parameters are chosen to be the largest ones. With the same autoencoder as LDMVFI~\cite{danier2023ldmvfi}, our method (ours$\dagger$) achieves better performance than LDMVFI, indicating the effectiveness of our consecutive Brownian Bridge Diffusion. Moreover, with an improved autoencoder, our method (denoted as ours) achieves state-of-the-art performance. It is important to note that we achieve much better FloLPIPS than other SOTAs, indicating our interpolated results achieve stronger motion consistency.

\begin{figure}[t]
    \centering
    \includegraphics[width=\linewidth]{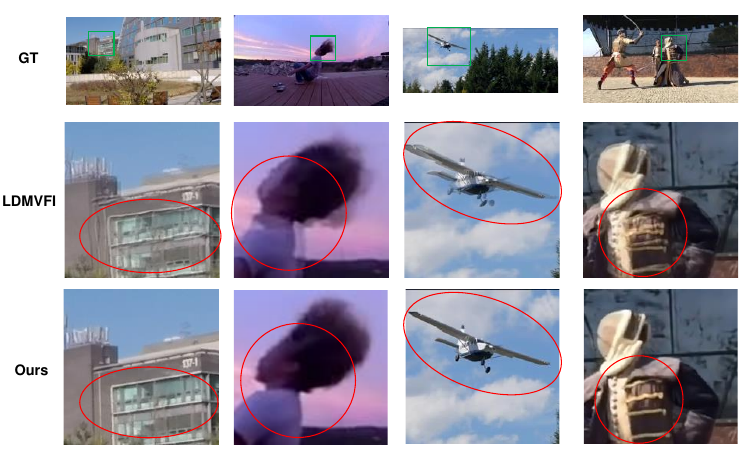}
    \vspace{1mm}
    \caption{\textbf{The reconstruction quality of our autoencoder and LDMVFI's autoencoder (decoding with ground truth latent representation x)}. Images are cropped with green boxes for detailed comparisons. Red circles highlight the details where our method achieves better performance. LDMVFI usually outputs overlaid images while our method does not.\vspace{-3mm}}
    \label{fig:compare}
\end{figure}

\begin{figure}[t]
    \centering
    \includegraphics[width=\linewidth]{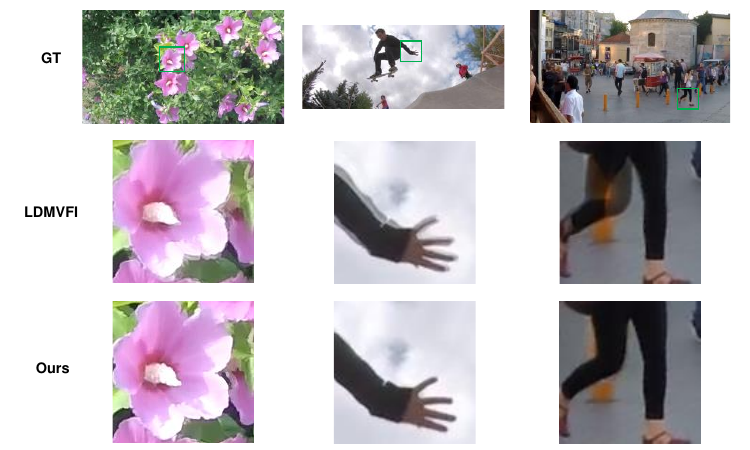}\vspace{5mm}
    \caption{The visual comparison of interpolated results of LDMVFI~\cite{danier2023ldmvfi} vs our method with the same autoencoder in LDMVFI (LDMVFI vs our$\dagger$ in Table~\ref{tab:results}). With the same autoencoder, our method can still achieve better visual quality than LDMVFI, demonstrating the superiority of our proposed consecutive Brownian Bridge diffusion.\vspace{5mm}}
    \label{fig:compare our daggar}
\end{figure}

\noindent\textbf{Qualitative Results.} In Table~\ref{tab:results}, our consecutive Brownian Bridge diffusion with the autoencoder in LDMVFI~\cite{danier2023ldmvfi} (denoted as our$\dagger$) generally achieves better quantitative results than LDMVFI, showing our method is effective. We include qualitative visualization in Figure~\ref{fig:compare our daggar} to support this result. Moreover, as mentioned in Section~\ref{sec:intro}, we find that the autoencoder in~\cite{danier2023ldmvfi} usually reconstructs overlaid images, and therefore we propose a new method of reconstruction. We provide examples to visualize the reconstruction results with our autoencoder and LDMVFI's autoencoder for comparison, shown in Figure~\ref{fig:compare}. All examples are from SNU-FILM extreme~\cite{choi2020channel}, which contains relatively large motion changes in neighboring frames.

We have provided some visual comparisons of our method and recent SOTAs in Figure~\ref{fig:teaser}. Our method achieves better visual quality because we have clearer details such as dog skins, cloth with folds, and fences with nets. However, UPR-Net~\cite{jin2023unified} achieves better PSNR/SSIM in all the cropped regions ($5-10\%$ better) than ours, which is highly inconsistent with the visual quality.

\subsection{Ablation Studies}
As we discussed in Section~\ref{sec: formulation}, latent-diffusion-based VFI is broken down into two stages, so we have a novel method to analyze the entire model. We conduct an ablation study on the ground truth estimation capability of our consecutive Brownian Bridge diffusion. We compare the evaluation results of decoded images with diffusion-generated latent representation $\hat{\mathbf{x}}$ and ground truth $\mathbf{x}$, which is encoded $I_n$. The results are shown in Table~\ref{tab:ablation}. It is important to note that, fixing inputs as the ground truth, our autoencoder achieves a stronger performance than the autoencoder in LDMVFI~\cite{danier2023ldmvfi}, indicating the effectiveness of our autoencoder. Also, fixing the autoencoder, our consecutive Brownian Bridge diffusion achieves almost identical performance with the ground truth, indicating its strong capability of ground truth estimation. However, the conditional generation model in LDMVFI~\cite{danier2023ldmvfi} usually underperforms the autoencoder with ground truth inputs. Therefore, our method has a stronger ability in both the autoencoder and ground truth estimation stages. More ablation studies are provided
in \textcolor{blue}{Appendix~\ref{appendix:ablation}}.

\section{Conclusion}
In this study, we propose our consecutive Brownian Bridge diffusion Model that better estimates the ground truth latent representation due to its low cumulative variance. We justify its effectiveness with extensive experiments in a wide range of datasets, though we do acknowledge that it requires larger GPU memory (18.55G for one $1080\times720$ image) than recent non-diffusion VFI methods such as UPR-Net~\cite{jin2023unified}(3G). Our method improves when the autoencoder is improved and achieves state-of-the-art performance with a simple yet effective design of the autoencoder, demonstrating its strong potential in the VFI task as a carefully designed autoencoder could potentially boost the performance by a large margin. In addition, we propose a novel method to analyze LDM-based VFI, providing insights for future research: whether future research could be conducted on autoencoder or diffusion model. Therefore, we believe our work will provide unique research directions and insights for diffusion-based video frame interpolation.

%\begin{acks}
%To Robert, for the bagels and explaining CMYK and color spaces.
%\end{acks}

\bibliographystyle{ACM-Reference-Format}

\bibliography{reference}


\begin{thebibliography}{60}

%%% the Web address.

\ifx \showCODEN    \undefined \def \showCODEN     #1{\unskip}     \fi
\ifx \showDOI      \undefined \def \showDOI       #1{#1}\fi
\ifx \showISBNx    \undefined \def \showISBNx     #1{\unskip}     \fi
\ifx \showISBNxiii \undefined \def \showISBNxiii  #1{\unskip}     \fi
\ifx \showISSN     \undefined \def \showISSN      #1{\unskip}     \fi
\ifx \showLCCN     \undefined \def \showLCCN      #1{\unskip}     \fi
\ifx \shownote     \undefined \def \shownote      #1{#1}          \fi
\ifx \showarticletitle \undefined \def \showarticletitle #1{#1}   \fi
\ifx \showURL      \undefined \def \showURL       {\relax}        \fi
% invisible to TeX
\providecommand\bibfield[2]{#2}
\providecommand\bibinfo[2]{#2}
\providecommand\natexlab[1]{#1}
\providecommand\showeprint[2][]{arXiv:#2}

\bibitem[Argaw and Kweon(2022)]%
        {argaw2022long}
\bibfield{author}{\bibinfo{person}{Dawit~Mureja Argaw} {and} \bibinfo{person}{In~So Kweon}.} \bibinfo{year}{2022}\natexlab{}.
\newblock \showarticletitle{Long-term video frame interpolation via feature propagation}. In \bibinfo{booktitle}{\emph{Proceedings of the IEEE/CVF Conference on Computer Vision and Pattern Recognition}}.
\newblock


\bibitem[Baker et~al\mbox{.}(2011)]%
        {baker2011database}
\bibfield{author}{\bibinfo{person}{Simon Baker}, \bibinfo{person}{Daniel Scharstein}, \bibinfo{person}{James~P Lewis}, \bibinfo{person}{Stefan Roth}, \bibinfo{person}{Michael~J Black}, {and} \bibinfo{person}{Richard Szeliski}.} \bibinfo{year}{2011}\natexlab{}.
\newblock \showarticletitle{A database and evaluation methodology for optical flow}.
\newblock \bibinfo{journal}{\emph{International journal of computer vision}} (\bibinfo{year}{2011}).
\newblock


\bibitem[Batzolis et~al\mbox{.}(2021)]%
        {batzolis2021conditional}
\bibfield{author}{\bibinfo{person}{Georgios Batzolis}, \bibinfo{person}{Jan Stanczuk}, \bibinfo{person}{Carola-Bibiane Sch{\"o}nlieb}, {and} \bibinfo{person}{Christian Etmann}.} \bibinfo{year}{2021}\natexlab{}.
\newblock \showarticletitle{Conditional image generation with score-based diffusion models}.
\newblock \bibinfo{journal}{\emph{arXiv preprint arXiv:2111.13606}} (\bibinfo{year}{2021}).
\newblock


\bibitem[Chen et~al\mbox{.}(2021)]%
        {chen2021pdwn}
\bibfield{author}{\bibinfo{person}{Zhiqi Chen}, \bibinfo{person}{Ran Wang}, \bibinfo{person}{Haojie Liu}, {and} \bibinfo{person}{Yao Wang}.} \bibinfo{year}{2021}\natexlab{}.
\newblock \showarticletitle{PDWN: Pyramid deformable warping network for video interpolation}.
\newblock \bibinfo{journal}{\emph{IEEE Open Journal of Signal Processing}} (\bibinfo{year}{2021}).
\newblock


\bibitem[Cheng and Chen(2020)]%
        {cheng2020video}
\bibfield{author}{\bibinfo{person}{Xianhang Cheng} {and} \bibinfo{person}{Zhenzhong Chen}.} \bibinfo{year}{2020}\natexlab{}.
\newblock \showarticletitle{Video frame interpolation via deformable separable convolution}. In \bibinfo{booktitle}{\emph{Proceedings of the AAAI Conference on Artificial Intelligence}}.
\newblock


\bibitem[Choi et~al\mbox{.}(2021)]%
        {choi2021high}
\bibfield{author}{\bibinfo{person}{Jinsoo Choi}, \bibinfo{person}{Jaesik Park}, {and} \bibinfo{person}{In~So Kweon}.} \bibinfo{year}{2021}\natexlab{}.
\newblock \showarticletitle{High-quality frame interpolation via tridirectional inference}. In \bibinfo{booktitle}{\emph{Proceedings of the IEEE/CVF Winter Conference on Applications of Computer Vision}}.
\newblock


\bibitem[Choi et~al\mbox{.}(2020)]%
        {choi2020channel}
\bibfield{author}{\bibinfo{person}{Myungsub Choi}, \bibinfo{person}{Heewon Kim}, \bibinfo{person}{Bohyung Han}, \bibinfo{person}{Ning Xu}, {and} \bibinfo{person}{Kyoung~Mu Lee}.} \bibinfo{year}{2020}\natexlab{}.
\newblock \showarticletitle{Channel attention is all you need for video frame interpolation}. In \bibinfo{booktitle}{\emph{Proceedings of the AAAI Conference on Artificial Intelligence}}.
\newblock


\bibitem[Dai et~al\mbox{.}(2017)]%
        {dai2017deformable}
\bibfield{author}{\bibinfo{person}{Jifeng Dai}, \bibinfo{person}{Haozhi Qi}, \bibinfo{person}{Yuwen Xiong}, \bibinfo{person}{Yi Li}, \bibinfo{person}{Guodong Zhang}, \bibinfo{person}{Han Hu}, {and} \bibinfo{person}{Yichen Wei}.} \bibinfo{year}{2017}\natexlab{}.
\newblock \showarticletitle{Deformable convolutional networks}. In \bibinfo{booktitle}{\emph{Proceedings of the IEEE international conference on computer vision}}.
\newblock


\bibitem[Danier et~al\mbox{.}(2022)]%
        {danier2022flolpips}
\bibfield{author}{\bibinfo{person}{Duolikun Danier}, \bibinfo{person}{Fan Zhang}, {and} \bibinfo{person}{David Bull}.} \bibinfo{year}{2022}\natexlab{}.
\newblock \showarticletitle{FloLPIPS: A bespoke video quality metric for frame interpolation}. In \bibinfo{booktitle}{\emph{2022 Picture Coding Symposium (PCS)}}. IEEE.
\newblock


\bibitem[Danier et~al\mbox{.}(2024)]%
        {danier2023ldmvfi}
\bibfield{author}{\bibinfo{person}{Duolikun Danier}, \bibinfo{person}{Fan Zhang}, {and} \bibinfo{person}{David~R. Bull}.} \bibinfo{year}{2024}\natexlab{}.
\newblock \showarticletitle{LDMVFI: Video Frame Interpolation with Latent Diffusion Models}. In \bibinfo{booktitle}{\emph{AAAI Conference on Artificial Intelligence}}.
\newblock


\bibitem[De~Bortoli et~al\mbox{.}(2021)]%
        {de2021diffusion}
\bibfield{author}{\bibinfo{person}{Valentin De~Bortoli}, \bibinfo{person}{James Thornton}, \bibinfo{person}{Jeremy Heng}, {and} \bibinfo{person}{Arnaud Doucet}.} \bibinfo{year}{2021}\natexlab{}.
\newblock \showarticletitle{Diffusion schr{\"o}dinger bridge with applications to score-based generative modeling}.
\newblock \bibinfo{journal}{\emph{Advances in Neural Information Processing Systems}} (\bibinfo{year}{2021}).
\newblock


\bibitem[Dutta et~al\mbox{.}(2022)]%
        {dutta2022non}
\bibfield{author}{\bibinfo{person}{Saikat Dutta}, \bibinfo{person}{Arulkumar Subramaniam}, {and} \bibinfo{person}{Anurag Mittal}.} \bibinfo{year}{2022}\natexlab{}.
\newblock \showarticletitle{Non-linear motion estimation for video frame interpolation using space-time convolutions}. In \bibinfo{booktitle}{\emph{Proceedings of the IEEE/CVF conference on computer vision and pattern recognition}}.
\newblock


\bibitem[Flynn et~al\mbox{.}(2016)]%
        {flynn2016deepstereo}
\bibfield{author}{\bibinfo{person}{John Flynn}, \bibinfo{person}{Ivan Neulander}, \bibinfo{person}{James Philbin}, {and} \bibinfo{person}{Noah Snavely}.} \bibinfo{year}{2016}\natexlab{}.
\newblock \showarticletitle{Deepstereo: Learning to predict new views from the world's imagery}. In \bibinfo{booktitle}{\emph{Proceedings of the IEEE conference on computer vision and pattern recognition}}.
\newblock


\bibitem[Hang et~al\mbox{.}(2023)]%
        {hang2023efficient}
\bibfield{author}{\bibinfo{person}{Tiankai Hang}, \bibinfo{person}{Shuyang Gu}, \bibinfo{person}{Chen Li}, \bibinfo{person}{Jianmin Bao}, \bibinfo{person}{Dong Chen}, \bibinfo{person}{Han Hu}, \bibinfo{person}{Xin Geng}, {and} \bibinfo{person}{Baining Guo}.} \bibinfo{year}{2023}\natexlab{}.
\newblock \showarticletitle{Efficient diffusion training via min-snr weighting strategy}. In \bibinfo{booktitle}{\emph{Proceedings of the IEEE/CVF International Conference on Computer Vision}}.
\newblock


\bibitem[Heusel et~al\mbox{.}(2017)]%
        {heusel2017gans}
\bibfield{author}{\bibinfo{person}{Martin Heusel}, \bibinfo{person}{Hubert Ramsauer}, \bibinfo{person}{Thomas Unterthiner}, \bibinfo{person}{Bernhard Nessler}, {and} \bibinfo{person}{Sepp Hochreiter}.} \bibinfo{year}{2017}\natexlab{}.
\newblock \showarticletitle{Gans trained by a two time-scale update rule converge to a local nash equilibrium}.
\newblock \bibinfo{journal}{\emph{Advances in neural information processing systems}} (\bibinfo{year}{2017}).
\newblock


\bibitem[Ho et~al\mbox{.}(2020)]%
        {ho2020denoising}
\bibfield{author}{\bibinfo{person}{Jonathan Ho}, \bibinfo{person}{Ajay Jain}, {and} \bibinfo{person}{Pieter Abbeel}.} \bibinfo{year}{2020}\natexlab{}.
\newblock \showarticletitle{Denoising diffusion probabilistic models}.
\newblock \bibinfo{journal}{\emph{Advances in neural information processing systems}} (\bibinfo{year}{2020}).
\newblock


\bibitem[Hu et~al\mbox{.}(2022)]%
        {hu2022many}
\bibfield{author}{\bibinfo{person}{Ping Hu}, \bibinfo{person}{Simon Niklaus}, \bibinfo{person}{Stan Sclaroff}, {and} \bibinfo{person}{Kate Saenko}.} \bibinfo{year}{2022}\natexlab{}.
\newblock \showarticletitle{Many-to-many splatting for efficient video frame interpolation}. In \bibinfo{booktitle}{\emph{Proceedings of the IEEE/CVF Conference on Computer Vision and Pattern Recognition}}.
\newblock


\bibitem[Huang et~al\mbox{.}(2022a)]%
        {huang2022flowformer}
\bibfield{author}{\bibinfo{person}{Zhaoyang Huang}, \bibinfo{person}{Xiaoyu Shi}, \bibinfo{person}{Chao Zhang}, \bibinfo{person}{Qiang Wang}, \bibinfo{person}{Ka~Chun Cheung}, \bibinfo{person}{Hongwei Qin}, \bibinfo{person}{Jifeng Dai}, {and} \bibinfo{person}{Hongsheng Li}.} \bibinfo{year}{2022}\natexlab{a}.
\newblock \showarticletitle{Flowformer: A transformer architecture for optical flow}. In \bibinfo{booktitle}{\emph{European conference on computer vision}}.
\newblock


\bibitem[Huang et~al\mbox{.}(2022b)]%
        {huang2022real}
\bibfield{author}{\bibinfo{person}{Zhewei Huang}, \bibinfo{person}{Tianyuan Zhang}, \bibinfo{person}{Wen Heng}, \bibinfo{person}{Boxin Shi}, {and} \bibinfo{person}{Shuchang Zhou}.} \bibinfo{year}{2022}\natexlab{b}.
\newblock \showarticletitle{Real-time intermediate flow estimation for video frame interpolation}. In \bibinfo{booktitle}{\emph{European Conference on Computer Vision}}.
\newblock


\bibitem[Huang et~al\mbox{.}(2022c)]%
        {huang2022rife}
\bibfield{author}{\bibinfo{person}{Zhewei Huang}, \bibinfo{person}{Tianyuan Zhang}, \bibinfo{person}{Wen Heng}, \bibinfo{person}{Boxin Shi}, {and} \bibinfo{person}{Shuchang Zhou}.} \bibinfo{year}{2022}\natexlab{c}.
\newblock \showarticletitle{Real-Time Intermediate Flow Estimation for Video Frame Interpolation}. In \bibinfo{booktitle}{\emph{Proceedings of the European Conference on Computer Vision (ECCV)}}.
\newblock


\bibitem[Hui et~al\mbox{.}(2018)]%
        {hui2018liteflownet}
\bibfield{author}{\bibinfo{person}{Tak-Wai Hui}, \bibinfo{person}{Xiaoou Tang}, {and} \bibinfo{person}{Chen~Change Loy}.} \bibinfo{year}{2018}\natexlab{}.
\newblock \showarticletitle{Liteflownet: A lightweight convolutional neural network for optical flow estimation}. In \bibinfo{booktitle}{\emph{Proceedings of the IEEE conference on computer vision and pattern recognition}}.
\newblock


\bibitem[Ilg et~al\mbox{.}(2017)]%
        {ilg2017flownet}
\bibfield{author}{\bibinfo{person}{Eddy Ilg}, \bibinfo{person}{Nikolaus Mayer}, \bibinfo{person}{Tonmoy Saikia}, \bibinfo{person}{Margret Keuper}, \bibinfo{person}{Alexey Dosovitskiy}, {and} \bibinfo{person}{Thomas Brox}.} \bibinfo{year}{2017}\natexlab{}.
\newblock \showarticletitle{Flownet 2.0: Evolution of optical flow estimation with deep networks}. In \bibinfo{booktitle}{\emph{Proceedings of the IEEE conference on computer vision and pattern recognition}}.
\newblock


\bibitem[Jin et~al\mbox{.}(2023)]%
        {jin2023unified}
\bibfield{author}{\bibinfo{person}{Xin Jin}, \bibinfo{person}{Longhai Wu}, \bibinfo{person}{Jie Chen}, \bibinfo{person}{Youxin Chen}, \bibinfo{person}{Jayoon Koo}, {and} \bibinfo{person}{Cheul-hee Hahm}.} \bibinfo{year}{2023}\natexlab{}.
\newblock \showarticletitle{A unified pyramid recurrent network for video frame interpolation}. In \bibinfo{booktitle}{\emph{Proceedings of the IEEE/CVF Conference on Computer Vision and Pattern Recognition}}.
\newblock


\bibitem[Kingma and Ba(2015)]%
        {Adam}
\bibfield{author}{\bibinfo{person}{Diederik~P. Kingma} {and} \bibinfo{person}{Jimmy Ba}.} \bibinfo{year}{2015}\natexlab{}.
\newblock \showarticletitle{Adam: {A} Method for Stochastic Optimization}. In \bibinfo{booktitle}{\emph{International Conference on Learning Representations}}.
\newblock


\bibitem[Kong et~al\mbox{.}(2022)]%
        {Kong_2022_CVPR}
\bibfield{author}{\bibinfo{person}{Lingtong Kong}, \bibinfo{person}{Boyuan Jiang}, \bibinfo{person}{Donghao Luo}, \bibinfo{person}{Wenqing Chu}, \bibinfo{person}{Xiaoming Huang}, \bibinfo{person}{Ying Tai}, \bibinfo{person}{Chengjie Wang}, {and} \bibinfo{person}{Jie Yang}.} \bibinfo{year}{2022}\natexlab{}.
\newblock \showarticletitle{IFRNet: Intermediate Feature Refine Network for Efficient Frame Interpolation}. In \bibinfo{booktitle}{\emph{Proceedings of the IEEE/CVF Conference on Computer Vision and Pattern Recognition (CVPR)}}.
\newblock


\bibitem[Lee et~al\mbox{.}(2020)]%
        {lee2020adacof}
\bibfield{author}{\bibinfo{person}{Hyeongmin Lee}, \bibinfo{person}{Taeoh Kim}, \bibinfo{person}{Tae-young Chung}, \bibinfo{person}{Daehyun Pak}, \bibinfo{person}{Yuseok Ban}, {and} \bibinfo{person}{Sangyoun Lee}.} \bibinfo{year}{2020}\natexlab{}.
\newblock \showarticletitle{Adacof: Adaptive collaboration of flows for video frame interpolation}. In \bibinfo{booktitle}{\emph{Proceedings of the IEEE/CVF conference on computer vision and pattern recognition}}.
\newblock


\bibitem[Li et~al\mbox{.}(2023a)]%
        {li2023bbdm}
\bibfield{author}{\bibinfo{person}{Bo Li}, \bibinfo{person}{Kaitao Xue}, \bibinfo{person}{Bin Liu}, {and} \bibinfo{person}{Yu-Kun Lai}.} \bibinfo{year}{2023}\natexlab{a}.
\newblock \showarticletitle{BBDM: Image-to-image translation with Brownian bridge diffusion models}. In \bibinfo{booktitle}{\emph{Proceedings of the IEEE/CVF Conference on Computer Vision and Pattern Recognition}}.
\newblock


\bibitem[Li et~al\mbox{.}(2023b)]%
        {licvpr23amt}
\bibfield{author}{\bibinfo{person}{Zhen Li}, \bibinfo{person}{Zuo-Liang Zhu}, \bibinfo{person}{Ling-Hao Han}, \bibinfo{person}{Qibin Hou}, \bibinfo{person}{Chun-Le Guo}, {and} \bibinfo{person}{Ming-Ming Cheng}.} \bibinfo{year}{2023}\natexlab{b}.
\newblock \showarticletitle{AMT: All-Pairs Multi-Field Transforms for Efficient Frame Interpolation}. In \bibinfo{booktitle}{\emph{IEEE Conference on Computer Vision and Pattern Recognition (CVPR)}}.
\newblock


\bibitem[Lu et~al\mbox{.}(2022b)]%
        {lu2022dpm}
\bibfield{author}{\bibinfo{person}{Cheng Lu}, \bibinfo{person}{Yuhao Zhou}, \bibinfo{person}{Fan Bao}, \bibinfo{person}{Jianfei Chen}, \bibinfo{person}{Chongxuan Li}, {and} \bibinfo{person}{Jun Zhu}.} \bibinfo{year}{2022}\natexlab{b}.
\newblock \showarticletitle{Dpm-solver: A fast ode solver for diffusion probabilistic model sampling in around 10 steps}.
\newblock \bibinfo{journal}{\emph{Advances in Neural Information Processing Systems}} (\bibinfo{year}{2022}).
\newblock


\bibitem[Lu et~al\mbox{.}(2022c)]%
        {lu2022dpm++}
\bibfield{author}{\bibinfo{person}{Cheng Lu}, \bibinfo{person}{Yuhao Zhou}, \bibinfo{person}{Fan Bao}, \bibinfo{person}{Jianfei Chen}, \bibinfo{person}{Chongxuan Li}, {and} \bibinfo{person}{Jun Zhu}.} \bibinfo{year}{2022}\natexlab{c}.
\newblock \showarticletitle{Dpm-solver++: Fast solver for guided sampling of diffusion probabilistic models}.
\newblock \bibinfo{journal}{\emph{arXiv preprint arXiv:2211.01095}} (\bibinfo{year}{2022}).
\newblock


\bibitem[Lu et~al\mbox{.}(2022a)]%
        {lu2022video}
\bibfield{author}{\bibinfo{person}{Liying Lu}, \bibinfo{person}{Ruizheng Wu}, \bibinfo{person}{Huaijia Lin}, \bibinfo{person}{Jiangbo Lu}, {and} \bibinfo{person}{Jiaya Jia}.} \bibinfo{year}{2022}\natexlab{a}.
\newblock \showarticletitle{Video frame interpolation with transformer}. In \bibinfo{booktitle}{\emph{Proceedings of the IEEE/CVF Conference on Computer Vision and Pattern Recognition}}.
\newblock


\bibitem[Niklaus and Liu(2018)]%
        {niklaus2018context}
\bibfield{author}{\bibinfo{person}{Simon Niklaus} {and} \bibinfo{person}{Feng Liu}.} \bibinfo{year}{2018}\natexlab{}.
\newblock \showarticletitle{Context-aware synthesis for video frame interpolation}. In \bibinfo{booktitle}{\emph{Proceedings of the IEEE conference on computer vision and pattern recognition}}.
\newblock


\bibitem[Niklaus and Liu(2020)]%
        {niklaus2020softmax}
\bibfield{author}{\bibinfo{person}{Simon Niklaus} {and} \bibinfo{person}{Feng Liu}.} \bibinfo{year}{2020}\natexlab{}.
\newblock \showarticletitle{Softmax splatting for video frame interpolation}. In \bibinfo{booktitle}{\emph{Proceedings of the IEEE/CVF conference on computer vision and pattern recognition}}.
\newblock


\bibitem[Niklaus et~al\mbox{.}(2017a)]%
        {niklaus2017video}
\bibfield{author}{\bibinfo{person}{Simon Niklaus}, \bibinfo{person}{Long Mai}, {and} \bibinfo{person}{Feng Liu}.} \bibinfo{year}{2017}\natexlab{a}.
\newblock \showarticletitle{Video frame interpolation via adaptive convolution}. In \bibinfo{booktitle}{\emph{Proceedings of the IEEE conference on computer vision and pattern recognition}}.
\newblock


\bibitem[Niklaus et~al\mbox{.}(2017b)]%
        {niklaus2017videosep}
\bibfield{author}{\bibinfo{person}{Simon Niklaus}, \bibinfo{person}{Long Mai}, {and} \bibinfo{person}{Feng Liu}.} \bibinfo{year}{2017}\natexlab{b}.
\newblock \showarticletitle{Video frame interpolation via adaptive separable convolution}. In \bibinfo{booktitle}{\emph{Proceedings of the IEEE international conference on computer vision}}.
\newblock


\bibitem[Oksendal(2013)]%
        {oksendal2013stochastic}
\bibfield{author}{\bibinfo{person}{Bernt Oksendal}.} \bibinfo{year}{2013}\natexlab{}.
\newblock \bibinfo{booktitle}{\emph{Stochastic differential equations: an introduction with applications}}.
\newblock \bibinfo{publisher}{Springer Science \& Business Media}.
\newblock


\bibitem[Park et~al\mbox{.}(2023)]%
        {park2023BiFormer}
\bibfield{author}{\bibinfo{person}{Junheum Park}, \bibinfo{person}{Jintae Kim}, {and} \bibinfo{person}{Chang-Su Kim}.} \bibinfo{year}{2023}\natexlab{}.
\newblock \showarticletitle{BiFormer: Learning Bilateral Motion Estimation via Bilateral Transformer for 4K Video Frame Interpolation}. In \bibinfo{booktitle}{\emph{Computer Vision and Pattern Recognition}}.
\newblock


\bibitem[Park et~al\mbox{.}(2021)]%
        {park2021ABME}
\bibfield{author}{\bibinfo{person}{Junheum Park}, \bibinfo{person}{Chul Lee}, {and} \bibinfo{person}{Chang-Su Kim}.} \bibinfo{year}{2021}\natexlab{}.
\newblock \showarticletitle{Asymmetric Bilateral Motion Estimation for Video Frame Interpolation}. In \bibinfo{booktitle}{\emph{International Conference on Computer Vision}}.
\newblock


\bibitem[Perazzi et~al\mbox{.}(2016)]%
        {Perazzi_CVPR_2016}
\bibfield{author}{\bibinfo{person}{Federico Perazzi}, \bibinfo{person}{Jordi Pont-Tuset}, \bibinfo{person}{Brian McWilliams}, \bibinfo{person}{Luc~Van Gool}, \bibinfo{person}{Markus Gross}, {and} \bibinfo{person}{Alexander Sorkine-Hornung}.} \bibinfo{year}{2016}\natexlab{}.
\newblock \showarticletitle{A Benchmark Dataset and Evaluation Methodology for Video Object Segmentation}. In \bibinfo{booktitle}{\emph{The IEEE Conference on Computer Vision and Pattern Recognition (CVPR)}}.
\newblock


\bibitem[Plack et~al\mbox{.}(2023)]%
        {plack2023frame}
\bibfield{author}{\bibinfo{person}{Markus Plack}, \bibinfo{person}{Karlis~Martins Briedis}, \bibinfo{person}{Abdelaziz Djelouah}, \bibinfo{person}{Matthias~B Hullin}, \bibinfo{person}{Markus Gross}, {and} \bibinfo{person}{Christopher Schroers}.} \bibinfo{year}{2023}\natexlab{}.
\newblock \showarticletitle{Frame Interpolation Transformer and Uncertainty Guidance}. In \bibinfo{booktitle}{\emph{Proceedings of the IEEE/CVF Conference on Computer Vision and Pattern Recognition}}.
\newblock


\bibitem[Rombach et~al\mbox{.}(2022)]%
        {rombach2022high}
\bibfield{author}{\bibinfo{person}{Robin Rombach}, \bibinfo{person}{Andreas Blattmann}, \bibinfo{person}{Dominik Lorenz}, \bibinfo{person}{Patrick Esser}, {and} \bibinfo{person}{Bj{\"o}rn Ommer}.} \bibinfo{year}{2022}\natexlab{}.
\newblock \showarticletitle{High-resolution image synthesis with latent diffusion models}. In \bibinfo{booktitle}{\emph{Proceedings of the IEEE/CVF conference on computer vision and pattern recognition}}.
\newblock


\bibitem[Ross(1995)]%
        {ross1995stochastic}
\bibfield{author}{\bibinfo{person}{Sheldon~M Ross}.} \bibinfo{year}{1995}\natexlab{}.
\newblock \bibinfo{booktitle}{\emph{Stochastic processes}}.
\newblock


\bibitem[Shi et~al\mbox{.}(2024)]%
        {shi2024diffusion}
\bibfield{author}{\bibinfo{person}{Yuyang Shi}, \bibinfo{person}{Valentin De~Bortoli}, \bibinfo{person}{Andrew Campbell}, {and} \bibinfo{person}{Arnaud Doucet}.} \bibinfo{year}{2024}\natexlab{}.
\newblock \showarticletitle{Diffusion Schr{\"o}dinger bridge matching}.
\newblock \bibinfo{journal}{\emph{Advances in Neural Information Processing Systems}} (\bibinfo{year}{2024}).
\newblock


\bibitem[Shi et~al\mbox{.}(2021)]%
        {shi2021video}
\bibfield{author}{\bibinfo{person}{Zhihao Shi}, \bibinfo{person}{Xiaohong Liu}, \bibinfo{person}{Kangdi Shi}, \bibinfo{person}{Linhui Dai}, {and} \bibinfo{person}{Jun Chen}.} \bibinfo{year}{2021}\natexlab{}.
\newblock \showarticletitle{Video frame interpolation via generalized deformable convolution}.
\newblock \bibinfo{journal}{\emph{IEEE transactions on multimedia}} (\bibinfo{year}{2021}).
\newblock


\bibitem[Siyao et~al\mbox{.}(2021)]%
        {siyao2021deep}
\bibfield{author}{\bibinfo{person}{Li Siyao}, \bibinfo{person}{Shiyu Zhao}, \bibinfo{person}{Weijiang Yu}, \bibinfo{person}{Wenxiu Sun}, \bibinfo{person}{Dimitris Metaxas}, \bibinfo{person}{Chen~Change Loy}, {and} \bibinfo{person}{Ziwei Liu}.} \bibinfo{year}{2021}\natexlab{}.
\newblock \showarticletitle{Deep animation video interpolation in the wild}. In \bibinfo{booktitle}{\emph{Proceedings of the IEEE/CVF conference on computer vision and pattern recognition}}.
\newblock


\bibitem[Song et~al\mbox{.}(2021a)]%
        {song2021denoising}
\bibfield{author}{\bibinfo{person}{Jiaming Song}, \bibinfo{person}{Chenlin Meng}, {and} \bibinfo{person}{Stefano Ermon}.} \bibinfo{year}{2021}\natexlab{a}.
\newblock \showarticletitle{Denoising Diffusion Implicit Models}. In \bibinfo{booktitle}{\emph{International Conference on Learning Representations}}.
\newblock


\bibitem[Song et~al\mbox{.}(2021b)]%
        {song2021scorebased}
\bibfield{author}{\bibinfo{person}{Yang Song}, \bibinfo{person}{Jascha Sohl-Dickstein}, \bibinfo{person}{Diederik~P Kingma}, \bibinfo{person}{Abhishek Kumar}, \bibinfo{person}{Stefano Ermon}, {and} \bibinfo{person}{Ben Poole}.} \bibinfo{year}{2021}\natexlab{b}.
\newblock \showarticletitle{Score-Based Generative Modeling through Stochastic Differential Equations}. In \bibinfo{booktitle}{\emph{International Conference on Learning Representations}}.
\newblock


\bibitem[Soomro et~al\mbox{.}(2012)]%
        {soomro2012ucf101}
\bibfield{author}{\bibinfo{person}{Khurram Soomro}, \bibinfo{person}{Amir~Roshan Zamir}, {and} \bibinfo{person}{Mubarak Shah}.} \bibinfo{year}{2012}\natexlab{}.
\newblock \showarticletitle{UCF101: A dataset of 101 human actions classes from videos in the wild}.
\newblock \bibinfo{journal}{\emph{arXiv preprint arXiv:1212.0402}} (\bibinfo{year}{2012}).
\newblock


\bibitem[Sun et~al\mbox{.}(2018)]%
        {sun2018pwc}
\bibfield{author}{\bibinfo{person}{Deqing Sun}, \bibinfo{person}{Xiaodong Yang}, \bibinfo{person}{Ming-Yu Liu}, {and} \bibinfo{person}{Jan Kautz}.} \bibinfo{year}{2018}\natexlab{}.
\newblock \showarticletitle{Pwc-net: Cnns for optical flow using pyramid, warping, and cost volume}. In \bibinfo{booktitle}{\emph{Proceedings of the IEEE conference on computer vision and pattern recognition}}.
\newblock


\bibitem[Teed and Deng(2020)]%
        {teed2020raft}
\bibfield{author}{\bibinfo{person}{Zachary Teed} {and} \bibinfo{person}{Jia Deng}.} \bibinfo{year}{2020}\natexlab{}.
\newblock \showarticletitle{Raft: Recurrent all-pairs field transforms for optical flow}. In \bibinfo{booktitle}{\emph{European Conference on Computer Vision}}.
\newblock


\bibitem[Tu et~al\mbox{.}(2022)]%
        {tu2022maxvit}
\bibfield{author}{\bibinfo{person}{Zhengzhong Tu}, \bibinfo{person}{Hossein Talebi}, \bibinfo{person}{Han Zhang}, \bibinfo{person}{Feng Yang}, \bibinfo{person}{Peyman Milanfar}, \bibinfo{person}{Alan Bovik}, {and} \bibinfo{person}{Yinxiao Li}.} \bibinfo{year}{2022}\natexlab{}.
\newblock \showarticletitle{Maxvit: Multi-axis vision transformer}. In \bibinfo{booktitle}{\emph{European conference on computer vision}}.
\newblock


\bibitem[Vaswani et~al\mbox{.}(2017)]%
        {vaswani2017attention}
\bibfield{author}{\bibinfo{person}{Ashish Vaswani}, \bibinfo{person}{Noam Shazeer}, \bibinfo{person}{Niki Parmar}, \bibinfo{person}{Jakob Uszkoreit}, \bibinfo{person}{Llion Jones}, \bibinfo{person}{Aidan~N Gomez}, \bibinfo{person}{{\L}ukasz Kaiser}, {and} \bibinfo{person}{Illia Polosukhin}.} \bibinfo{year}{2017}\natexlab{}.
\newblock \showarticletitle{Attention is all you need}.
\newblock \bibinfo{journal}{\emph{Advances in neural information processing systems}} (\bibinfo{year}{2017}).
\newblock


\bibitem[Voleti et~al\mbox{.}(2022)]%
        {voleti2022mcvd}
\bibfield{author}{\bibinfo{person}{Vikram Voleti}, \bibinfo{person}{Alexia Jolicoeur-Martineau}, {and} \bibinfo{person}{Chris Pal}.} \bibinfo{year}{2022}\natexlab{}.
\newblock \showarticletitle{Mcvd-masked conditional video diffusion for prediction, generation, and interpolation}.
\newblock \bibinfo{journal}{\emph{Advances in neural information processing systems}} (\bibinfo{year}{2022}).
\newblock


\bibitem[Wang et~al\mbox{.}(2004)]%
        {wang2004image}
\bibfield{author}{\bibinfo{person}{Zhou Wang}, \bibinfo{person}{Alan~C Bovik}, \bibinfo{person}{Hamid~R Sheikh}, {and} \bibinfo{person}{Eero~P Simoncelli}.} \bibinfo{year}{2004}\natexlab{}.
\newblock \showarticletitle{Image quality assessment: from error visibility to structural similarity}.
\newblock \bibinfo{journal}{\emph{IEEE transactions on image processing}} (\bibinfo{year}{2004}).
\newblock


\bibitem[Weinzaepfel et~al\mbox{.}(2023)]%
        {weinzaepfel2023croco}
\bibfield{author}{\bibinfo{person}{Philippe Weinzaepfel}, \bibinfo{person}{Thomas Lucas}, \bibinfo{person}{Vincent Leroy}, \bibinfo{person}{Yohann Cabon}, \bibinfo{person}{Vaibhav Arora}, \bibinfo{person}{Romain Br{\'e}gier}, \bibinfo{person}{Gabriela Csurka}, \bibinfo{person}{Leonid Antsfeld}, \bibinfo{person}{Boris Chidlovskii}, {and} \bibinfo{person}{J{\'e}r{\^o}me Revaud}.} \bibinfo{year}{2023}\natexlab{}.
\newblock \showarticletitle{CroCo v2: Improved Cross-view Completion Pre-training for Stereo Matching and Optical Flow}. In \bibinfo{booktitle}{\emph{Proceedings of the IEEE/CVF International Conference on Computer Vision}}.
\newblock


\bibitem[Wu et~al\mbox{.}(2018)]%
        {wu2018video}
\bibfield{author}{\bibinfo{person}{Chao-Yuan Wu}, \bibinfo{person}{Nayan Singhal}, {and} \bibinfo{person}{Philipp Krahenbuhl}.} \bibinfo{year}{2018}\natexlab{}.
\newblock \showarticletitle{Video compression through image interpolation}. In \bibinfo{booktitle}{\emph{Proceedings of the European conference on computer vision (ECCV)}}.
\newblock


\bibitem[Xue et~al\mbox{.}(2019)]%
        {xue2019video}
\bibfield{author}{\bibinfo{person}{Tianfan Xue}, \bibinfo{person}{Baian Chen}, \bibinfo{person}{Jiajun Wu}, \bibinfo{person}{Donglai Wei}, {and} \bibinfo{person}{William~T Freeman}.} \bibinfo{year}{2019}\natexlab{}.
\newblock \showarticletitle{Video Enhancement with Task-Oriented Flow}.
\newblock \bibinfo{journal}{\emph{International Journal of Computer Vision (IJCV)}} (\bibinfo{year}{2019}).
\newblock


\bibitem[Zhang et~al\mbox{.}(2023)]%
        {zhang2023extracting}
\bibfield{author}{\bibinfo{person}{Guozhen Zhang}, \bibinfo{person}{Yuhan Zhu}, \bibinfo{person}{Haonan Wang}, \bibinfo{person}{Youxin Chen}, \bibinfo{person}{Gangshan Wu}, {and} \bibinfo{person}{Limin Wang}.} \bibinfo{year}{2023}\natexlab{}.
\newblock \showarticletitle{Extracting motion and appearance via inter-frame attention for efficient video frame interpolation}. In \bibinfo{booktitle}{\emph{Proceedings of the IEEE/CVF Conference on Computer Vision and Pattern Recognition}}.
\newblock


\bibitem[Zhang et~al\mbox{.}(2018)]%
        {zhang2018perceptual}
\bibfield{author}{\bibinfo{person}{Richard Zhang}, \bibinfo{person}{Phillip Isola}, \bibinfo{person}{Alexei~A Efros}, \bibinfo{person}{Eli Shechtman}, {and} \bibinfo{person}{Oliver Wang}.} \bibinfo{year}{2018}\natexlab{}.
\newblock \showarticletitle{The Unreasonable Effectiveness of Deep Features as a Perceptual Metric}. In \bibinfo{booktitle}{\emph{CVPR}}.
\newblock


\bibitem[Zhou et~al\mbox{.}(2024)]%
        {zhou2024denoising}
\bibfield{author}{\bibinfo{person}{Linqi Zhou}, \bibinfo{person}{Aaron Lou}, \bibinfo{person}{Samar Khanna}, {and} \bibinfo{person}{Stefano Ermon}.} \bibinfo{year}{2024}\natexlab{}.
\newblock \showarticletitle{Denoising Diffusion Bridge Models}. In \bibinfo{booktitle}{\emph{The Twelfth International Conference on Learning Representations}}.
\newblock


\end{thebibliography}

\clearpage
\appendix

\section{Formula Derivation}
\label{tech_details}
\subsection{Consecutive Brownian Bridge}
\label{appendix:formula}
For $0<t<h$, if we have $s>t$, then the Markov property of the Wiener process produces:
\begin{equation*}
    W_s|(W_0,W_t,W_h) = W_s|(W_t,W_h)
\end{equation*}
Applying in our setting, this becomes: $W_t|W_T = \mathbf{x},W_{2T} = \mathbf{z}$ for $t>T$. Note that only the variance of the Wiener process is related to time, and the variance of general Brownian Bridge $W_t|(W_{t_1},W_{t_2})$ is $\frac{(t_2-t)(t-t_1)}{t_2-t_1}$. If we add any value simultaneously to $t_1,t_2,t$, the variance is unchanged. Therefore, we can subtract T in time to get $W_{s}|W_0 = \mathbf{x} ,W_{T} = \mathbf{z}$, where $s = t-T$.

If we have $s<t$, then it is important to know that $tW_{t^{-1}}$ is a Wiener process with the same distribution with $W_t$~\cite{oksendal2013stochastic}. We can add a small $\epsilon$ to time and use such transformation to obtain: 
\begin{equation*}
    \begin{aligned}
    &W_{s}|(W_0,W_{t},W_{h})&\\ = \text{ }&W_{s+\epsilon}|(W_\epsilon,W_{t+\epsilon},W_{h+\epsilon})& \\= \text{ }& (s+\epsilon)W_{(s+\epsilon)^{-1}}|\epsilon W_{\epsilon^{-1}},(t+\epsilon)W_{(t+\epsilon)^{-1}},(h+\epsilon)W_{(h+\epsilon)^{-1}}&\\
    = \text{ }& (s+\epsilon)W_{(s+\epsilon)^{-1}}|\epsilon W_{\epsilon^{-1}},(t+\epsilon)W_{(t+\epsilon)^{-1}} &\\
    =\text{ } &W_s|(W_0,W_t)&
    \end{aligned}
\end{equation*} 
In our method, this becomes $W_t|W_0 = \mathbf{y},W_T = \mathbf{x}$. The distribution is $\mathcal{N}(\frac{t}{T}\mathbf{y} + (1-\frac{t}{T}\mathbf{x}),\frac{t(T-t)}{T}\mathbf{I})$. Now, let's consider another process defined as $W_s|W_0 = \mathbf{x},W_T = \mathbf{y}$. The distribution is easy to derive: $\mathcal{N}(\frac{s}{T}\mathbf{x} + (1-\frac{s}{T}\mathbf{y}),\frac{s(T-s)}{T}\mathbf{I})$. With simple algebra, we can find that when $s = T-t$, the two distributions are equal. Thus, we finish the derivation of the distribution of consecutive Brownian Bridge.

\subsection{Cumulative Variance}
\label{appendix: cum var}
We denote $\mathbf{z}$ as standard Gaussian distribution. In DDPM~\cite{ho2020denoising}, $\mathbf{x}_{t-1} = \frac{1}{\sqrt{1-\beta_t}}\left(\mathbf{x}_t - \frac{\beta_t}{\sqrt{1-\alpha_t} }\epsilon_\theta\right) + \sqrt{\hat{\beta}_t}\mathbf{z}$. At the first step of generation, since $\mathbf{x}_{T}\sim \mathcal{N}(\mathbf{0},\mathbf{I})$ and $0<\beta_t<1$, we have:
\begin{equation*}
    \begin{aligned}
        Var(\mathbf{x}_{T-1}) &= Var\left(\frac{1}{\sqrt{1-\beta_t}}\left(\mathbf{x}_T - \frac{\beta_t}{\sqrt{1-\alpha_t} }\epsilon_\theta\right) + \sqrt{\hat{\beta}_t}\mathbf{z}\right)&\\
        &> Var\left(\frac{1}{\sqrt{1-\beta_t}}\mathbf{x}_T + \sqrt{\hat{\beta}_t}\mathbf{z}\right)&\\
        &> 1 + \hat{\beta}_t&
    \end{aligned}
\end{equation*}
Since $\epsilon_\theta$ takes random input, it has a positive variance. The following sampling steps have fixed inputs $x_t$, so the variance only contains $\hat{\beta}_t$. Therefore, the cumulative variance is larger than $1+\sum_t\hat{\beta}_t$, corresponding to \textbf{11.036} in real experiments. However, in our method, we have $\mathbf{x}_{t-\Delta_t} = \mathbf{x}_t - \frac{\Delta_t}{t}\epsilon_\theta + \sqrt{\frac{(t-\Delta_t)\Delta_t}{t}}\mathbf{z}$, and $\mathbf{x}_T$ is deterministic, we have:
\begin{equation*}
    \begin{aligned}
        Var(x_{t-\Delta_t}) &= Var\left(\mathbf{x}_t - \frac{\Delta_t}{t}\epsilon_\theta + \sqrt{\frac{(t-\Delta_t)\Delta_t}{t}}\mathbf{z}\right)&\\
        &= Var\left(\sqrt{\frac{(t-\Delta_t)\Delta_t}{t}}\mathbf{z}\right)&\\
        &< \Delta_t&
    \end{aligned}
\end{equation*}
Since $\epsilon_\theta$ takes fixed inputs, it has no variance. The cumulative variance is smaller than $\sum_t\Delta_t = T$, corresponding to \textbf{2} in our experiments. We mentioned this result in Section 3.4 in our main paper.

\section{Connection with Diffusion SDEs}
\label{SDE}
Our method can be easily written in score-based SDE~\cite{batzolis2021conditional,song2021scorebased,zhou2024denoising}. The forward process of score-based SDEs is defined as:
\begin{equation}
    \label{dq:diffusion SDE forward}
    d\mathbf{x} = f(x,t)dt + g(t)d\mathbf{w}.
\end{equation}
$f(x,t)$ is the drift term, and $g(t)$ is the dispersion term. $\mathbf{w}$ denotes the standard Wiener process. The corresponding reversed SDE is defined as:

\begin{equation}
    \label{dq:diffusion SDE reverse}
    d\mathbf{x} = \left[f(x,t) - g(t)^2\nabla_{\mathbf{x}}logp_t(\mathbf{x}) \right]dt + g(t)d\bar{\mathbf{w}}.
\end{equation}
The conditional generation counterpart is defined as:

\begin{equation}
    \label{dq:diffusion SDE condition}
    d\mathbf{x} = \left\{f(x,t) - g(t)^2\nabla_{\mathbf{x}}[logp_t(\mathbf{x}) +logp_t(\mathbf{y}|\mathbf{x}) ] \right\}dt + g(t)d\bar{\mathbf{w}}.
\end{equation}
The term $\mathbf{y}$ is the conditional control for generation. Moreover, there exists a deterministic ODE trajectory (probability flow ODE) with the same marginal distribution $p_t(x)$ with Eq.~\eqref{dq:diffusion SDE reverse}~\cite{song2021scorebased}:

\begin{equation}
    \label{dq:diffusion ODE reverse}
    d\mathbf{x} = \left[f(x,t) - \frac{1}{2}g(t)^2\nabla_{\mathbf{x}}logp_t(\mathbf{x}) \right]dt.
\end{equation}
Therefore, the it suffices to train a neural network $s_\theta$ estimating $\nabla_{\mathbf{x}}logp_t(\mathbf{x})$~\cite{song2021scorebased}.
 Indeed, Brownian Bridge can be written in SDE form by~\cite{oksendal2013stochastic}:
\begin{equation}
\label{eq: BB SDE forward}
    d\mathbf{x} = \frac{\mathbf{y}-\mathbf{x}_t}{T-t}dt + d\mathbf{w}.
\end{equation}
$\mathbf{y}$ is another endpoint of the Brownian Bridge. The reversed SDE is defined as:
\begin{equation}
\label{eq: BB SDE reverse}
    d\mathbf{x} = \left[\frac{\mathbf{y}-\mathbf{x}_t}{T-t} -\nabla_{\mathbf{x}}logp_t(\mathbf{x})\right]dt + d\bar{\mathbf{w}}.
\end{equation}
By our formulation, our proposed method is compatible with score-based SDEs. Moreover, compared with conditional SDEs in Eq.~\eqref{dq:diffusion SDE condition}, this formulation does not include $logp_t(\mathbf{y}|\mathbf{x})$ which needs estimation.

\section{Additional Results}
\label{results}

\subsection{Quantitative Results}
\label{appendix:psnr}
We provide the evaluation results in PSNR/SSIM in Table~\ref{tab:results PSNR}. Though our method does not have state-of-the-art (but still comparable with SOTAs) performance in PSNR/SSIM, it is due to the \textbf{inconsistency} between PSNR/SSIM and visual quality (see Section~\ref{appendix:qualitative} and Figure~\ref{fig:inconsistent}). Therefore, we choose LPIPS/FloLPIPS/FID as our main evaluation metrics. 
\subsection{Qualitative Reults}
\label{appendix:qualitative}
\begin{figure}[t]
    \centering
    \includegraphics[width=\linewidth]{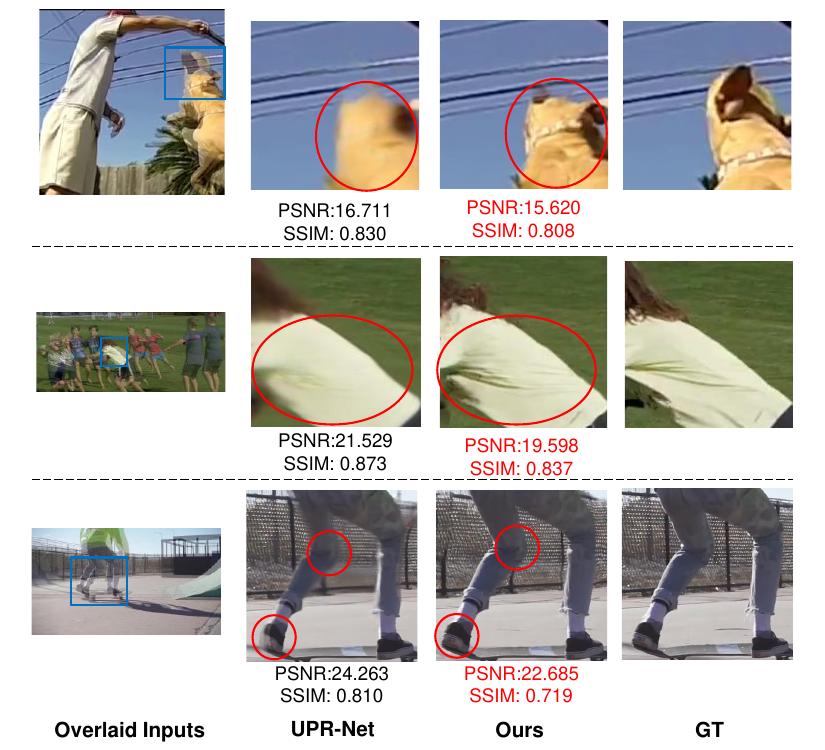}
    \caption{Visual illustration of the inconsistency between PSNR/SSIM and visual quality. Only images cropped within \textcolor{blue}{blue} boxes are evaluated with PSNR/SSIM. The red circles highlight our visual quality. Our method generates images with better visual quality, but the PSNR/SSIM is much lower.}
    \label{fig:inconsistent}
\end{figure}

\noindent\textbf{Inconsistency Between PSNR/SSIM and Visual Quality}. We provide some examples to demonstrate the inconsistency between PSNR/SSIM and visual quality, as shown in Figure~\ref{fig:inconsistent}. Our method achieves better visual quality than UPR-Net~\cite{jin2023unified} such as clearer dog skins, clearer cloth with folds, and clearer shoes and fences with nets. However, we did not achieve a satisfactory PSNR/SSIM, which is 5-10\% lower than that of UPR-Net. 

\begin{figure}[t]
    \centering
    \includegraphics[width=\linewidth]{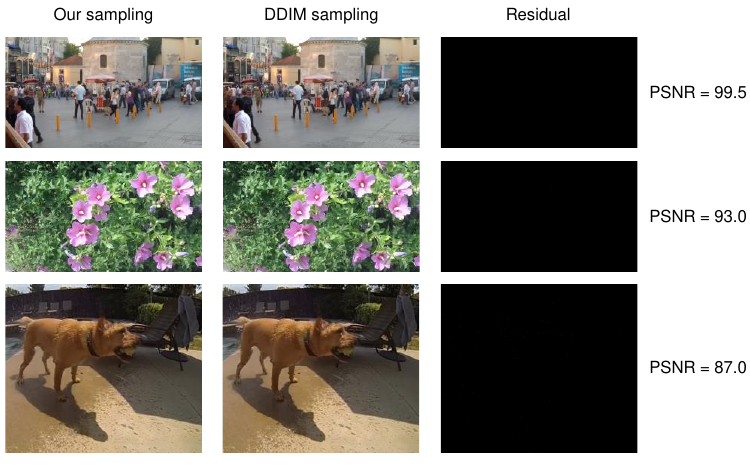}
    \caption{Visual comparison between our sampling and DDIM sampling with 5 steps generation. They achieve almost identical results (with very large PSNR). The residual is the absolute difference between the two images. Black means 0 difference, and almost everywhere is black.}
    \label{fig:vsddim}
\end{figure}

\begin{figure*}[ht]
    \centering
    \includegraphics[width=\linewidth]{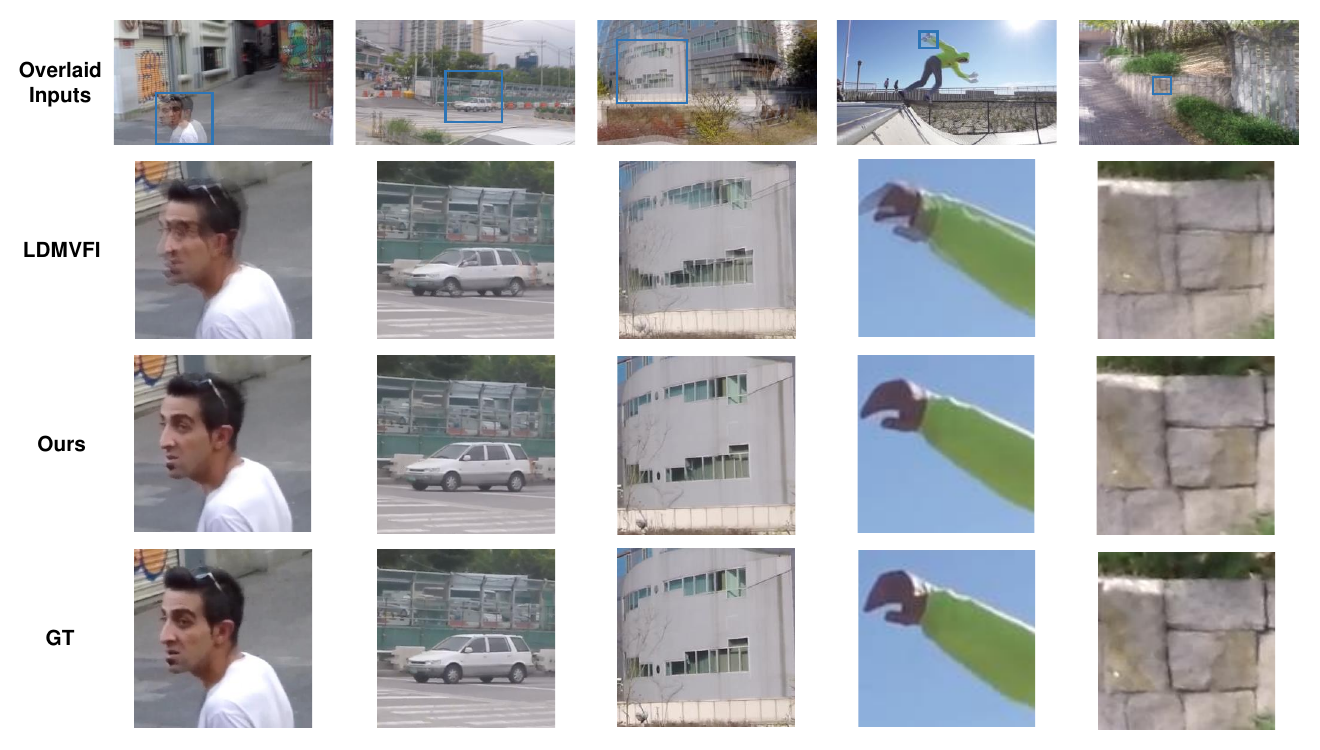}
    \caption{Additional Qualitative Comparison of our methods and LDMVFI. Images cropped with blue boxes are shown for better-detailed comparison. Our method steadily achieves better visual quality.}
    \label{fig: vsldm}
\end{figure*}

\begin{figure*}[ht]
    \centering
    \includegraphics[width=\linewidth]{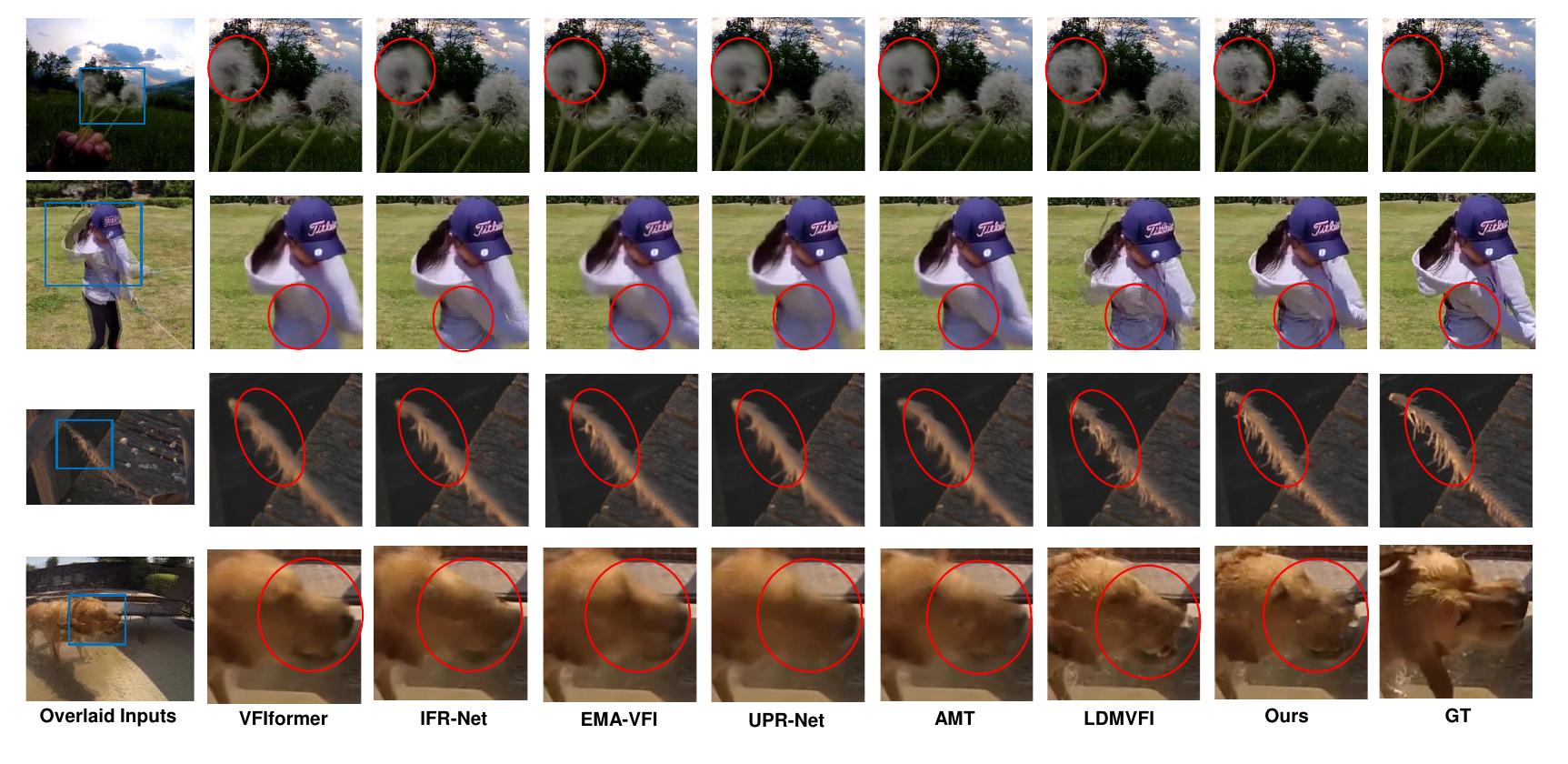}
    \caption{Additional Qualitative Comparison of our methods and recent SOTAs. Only images within the blue box are displayed for better-detailed comparison.}
    \label{fig:addtional qualitative}
\end{figure*}

\noindent\textbf{Additional Qualitative Comparisons.} In addition, we provide more qualitative comparisons between our method and LDMVFI~\cite{danier2023ldmvfi} in Figure~\ref{fig: vsldm} and qualitative comparisons between our method and recent SOTAs in Figure~\ref{fig:addtional qualitative}. All examples are selected from SNU-FILM extreme~\cite{choi2020channel}.

\begin{figure*}[ht]
    \centering
    \includegraphics[width=\linewidth]{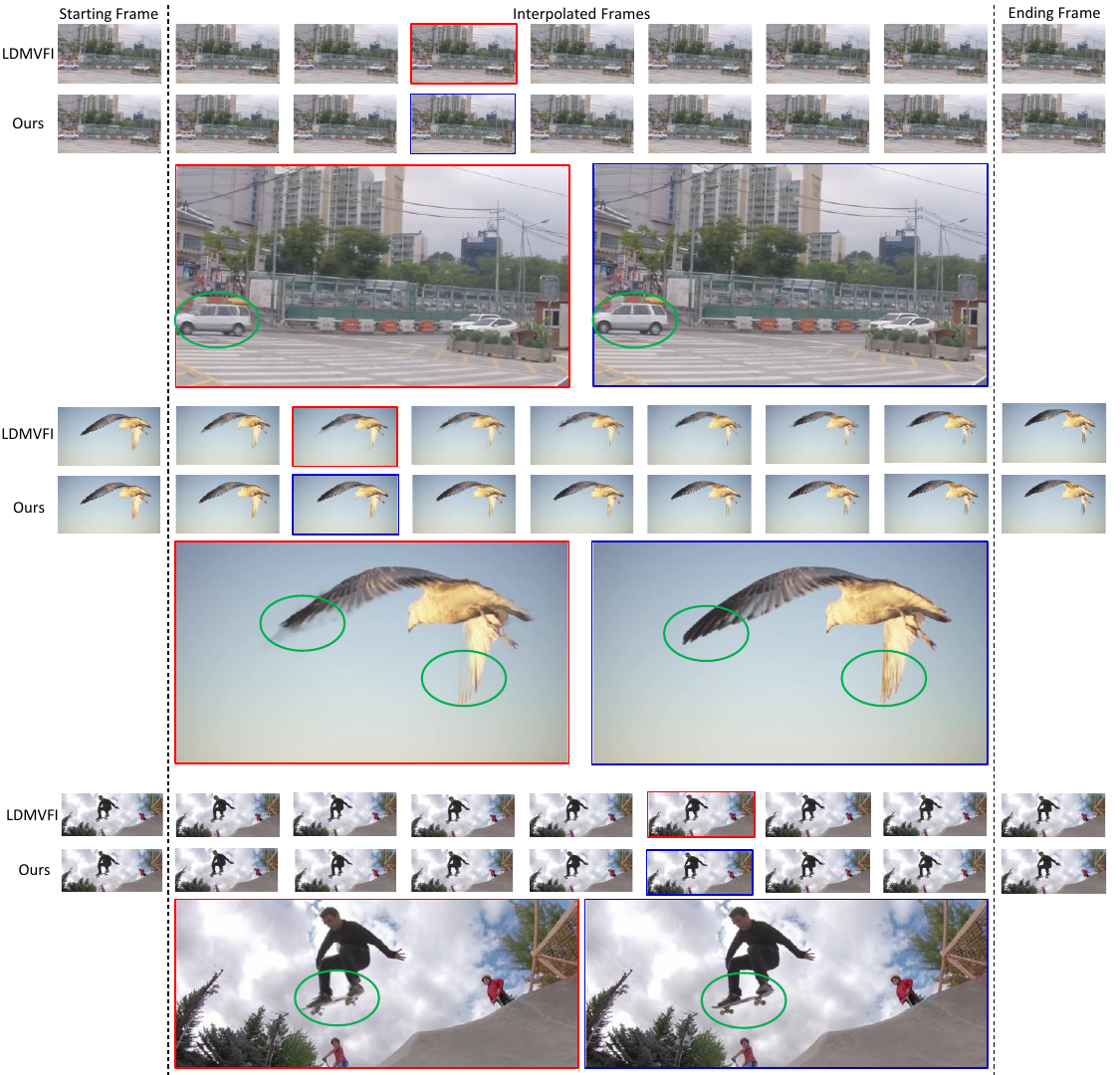}
    \caption{Multi-frame interpolation results. LDMVFI usually interpolates distorted or overlaid images while ours does not. Images with red and blue borders are displayed to show details. Our method corresponds to the blue border while LDMVFI corresponds to the red. Green circles highlight the detail where our performance is better.}
    \label{fig: interpolate}
\end{figure*}

\noindent\textbf{Multi-frame Interpolation.} We provide qualitative results of multi-frame interpolation of our methods and LDMVFI~\cite{danier2023ldmvfi}. Multi-frame interpolation is achieved in a bisection manner. We first interpolate $I_{0.5}$ with $I_0,I_1$, and then we interpolate $I_{0.25}$ with $I_0,I_{0.5}$ and $I_{0.75}$ with $I_{0.5},I_1$. More frames can be interpolated in this manner. We interpolate 7 frames between two $I_0,I_1$, and the visual comparisons are presented in Figure~\ref{fig: interpolate}. All examples are selected from SNU-FILM hard~\cite{choi2020channel}. Additional video demos are shown on our GitHub page: \href{https://zonglinl.github.io/videointerp}{\textcolor{blue}{https://zonglinl.github.io/videointerp}}. Due to the bisection-like multi-frame interpolation method, the multi-frame interpolation results largely depends on the first step of interpolation ($I_{0.5}$). If $I_{0.5}$ achieves good quality, then the relative motion in the second step (interpolating $I_{0.25},I_{0.75}$) is easy to achieve high quality because the motion changes become smaller. However, if the interpolation quality is not good at the first step, then later steps will not achieve good quality because such an unsatisfactory quality will be transmitted. LDMVFI~\ref{fig: vsldm} tends to generate overlaid or distorted $I_{0.5}$, resulting in unsatisfactory multi-frame interpolation results. \textcolor{blue}{We largely alleviate this problem, resulting in much better and more realistic interpolated videos.}

\noindent\textbf{Inference Time.} With one Nvidia RTX 8000 GPU, our method generates a $720\times1280$ image with approximately 1.2 seconds with 18G GPU memory. The inference speed is similar to recent SOTAs such as LDMVFI~\cite{danier2023ldmvfi} (1.2s) and UPR-Net-Large~\cite{jin2023unified} (1.15s), but diffusion-based methods require much more memory (LDMVFI requires 22G while UPR-Net requires 3G).

\begin{table}[t]
    \centering
    \caption{Ablation study on the number of sampling steps. This experiment is conducted on SNU-FILM extreme subset~\cite{choi2020channel}.}
    \label{tab:steps}
        \begin{tabular}{cccc}
        \toprule
        Number of steps &LPIPS & FloLPIPS& FID\\
        \midrule
        200 &0.110&0.184&36.632\\
        100 &0.110&0.184&36.631\\
        50 &0.110&0.184&36.631\\
        20 &0.110&0.184&36.632\\
        5 &0.110&0.184&36.632\\
        \bottomrule
    \end{tabular}
\end{table}
\subsection{Ablation Studies}
\label{appendix:ablation}

\textbf{Number of Sampling Steps.} We investigate how the number of sampling steps will impact the performance. This ablation study is conducted on SNU-FILM extreme subset~\cite{choi2020channel}, shown in Table~\ref{tab:steps}. We observe that the performance remains almost identical. The reason could be the relatively small differences between neighboring frames. Our method does not convert random noise to images like DDPM~\cite{ho2020denoising}. Instead, we convert one image to its neighboring frames, so we do not need to generate details from random noises. Instead, we change details from existing details, and therefore it may not need many steps to generate.

\noindent\textbf{DDIM Sampling.} As we claimed, our formulation does not need DDIM~\cite{song2021denoising} sampling to accelerate. We compare our sampling with DDIM sampling with $\eta = 0$ in 5 sampling steps for comparison. The visual result is shown in Figure~\ref{fig:vsddim}. There is almost no difference between the output of our sampling method and DDIM sampling, indicating that we do not require such a method to accelerate sampling.

\begin{table*}[t]
    \centering
    \caption{Quantitative results (PSNR/SSIM) on test datasets (the higher the better). $\dagger$ means we evaluate our consecutive Brownian Bridge diffusion (trained on Vimeo 90K~\cite{xue2019video}) with autoencoder provided by LDMVFI~\cite{danier2023ldmvfi}.}
    \label{tab:results PSNR}
    \resizebox{\textwidth}{!}{
        \begin{tabular}{cccccccc}
        \toprule
        \multirow{2}{*}{Methods}&\multirow{2}{*}{Middlebury} & \multirow{2}{*}{UCF-101}& \multirow{2}{*}{DAVIS}& \multicolumn{4}{c}{SNU-FILM} \\
        \cmidrule{5-8}
        & & & &easy & medium & hard&extreme\\
        &PSNR/SSIM&PSNR/SSIM&PSNR/SSIM&PSNR/SSIM&PSNR/SSIM&PSNR/SSIM&PSNR/SSIM\\
        \midrule
        ABME'21~\cite{park2021ABME} &37.639/0.986 &35.380/0.970 &26.861/0.865 & 39.590/0.990 & 35.770/0.979& 30.580/0.936&25.430/0.864\\
        
        MCVD'22~\cite{voleti2022mcvd} &20.539/0.820 &18.775/0.710& 18.946/0.705 &  22.201/0.828 & 21.488/0.812 &  20.314/0.766&18.464/0.694\\

        VFIformer'22~\cite{lu2022video} &  38.438/0.987 & 35.430/0.970 & 26.241/0.850 &40.130/0.991&36.090/0.980  & 30.670/0.938 & 25.430/0.864 \\
        
        IFRNet'22~\cite{Kong_2022_CVPR} & 36.368/0.983&35.420/0.967&27.313/0.877 &40.100/0.991&36.120/0.980 &30.630/0.937 &25.270/0.861 \\

        AMT'23~\cite{licvpr23amt} & 38.395/0.988 &35.450/0.970 & 27.234/0.877&39.880/0.991  &36.120/0.981&30.780/0.939 &25.430/0.865 \\
        UPR-Net'23~\cite{jin2023unified} &38.065/0.986&35.470/0.970& 26.894/0.870 & 40.440/0.991& 36.290/0.980&30.860/0.938& 25.630/0.864\\
        
        EMA-VFI'23~\cite{zhang2023extracting} &38.526/0.988 &35.480/0.970& 27.111/0.871&39.980/0.991&  36.090/0.980& 30.940/0.939&25.690/0.866 \\
        
        LDMVFI'24~\cite{danier2023ldmvfi} & 34.230/0.974&32.160/0.964&   25.073/0.819&38.890 0.988&33.975/0.971&29.144/0.911&23.349 0.827\\

        \midrule
        Ours$\dagger$ &34.057/0.970&34.730/0.965&25.446/0.837&38.720/0.988  &34.016/0.971&28.556/0.918 &23.931/0.837 \\
        
        Ours &36.852/0.983&35.151/0.968 &26.391/0.858 &39.637/0.990& 34.886/0.974& 29.615/0.929 &24.376/0.848 \\

        \bottomrule
    \end{tabular}
  }
\end{table*}

\clearpage
\end{document}